%% file: main.tex
\pdfoutput=1

\documentclass[11pt]{article}

\usepackage[final]{acl}

\usepackage{times}
\usepackage{latexsym}
\usepackage{inconsolata}
\usepackage{amsfonts,amssymb,amsmath}
\usepackage{multirow}
\usepackage{multicol}
\usepackage{hyperref}
\usepackage{booktabs}
\usepackage{threeparttable}
\usepackage{algorithm}
\usepackage{algorithmicx}
\usepackage{algpseudocode}
\usepackage{graphicx}
\usepackage{subfigure}
\usepackage[T1]{fontenc}

\usepackage[utf8]{inputenc}

\usepackage{microtype}

\usepackage{inconsolata}

\usepackage{graphicx}

\title{Enhancing Adversarial Transferability in Visual-Language Pre-training Models via Local Shuffle and Sample-based Attack}

\author{
  Xin Liu\thanks{The first two authors contributed equally.}, Aoyang Zhou\footnote[1]{}, Kun He\thanks{Corresponding author.} \\ 
  School of Computer Science and Technology \\ 
  Huazhong University of Science and Technology, Wuhan, China \\ 
  \texttt{liuxin\_jhl@hust.edu.cn},
  \texttt{aoyangzhou@hust.edu.cn},\texttt{brooklet60@hust.edu.cn}
}

\begin{document}
\maketitle
\begin{abstract}
\input{EMNLP2024/00abstract}
\end{abstract}

\section{Introduction}
\input{EMNLP2024/01Introduction}

\section{Related Work}
\input{EMNLP2024/02RW}

\section{Methodology}
\input{EMNLP2024/03Method}

\section{Experiments}
\input{EMNLP2024/04Experiments}

\section{Conclusion}
\input{EMNLP2024/05Conclusion}

\section*{Limitations}
LSSA focuses on the two popular and typical vision-language pre-training model in the multimodal task. However, it does not consider other types of modalities, such as video and signal. These data can also be crafted adversarial examples by applying our method. We will continue to explore the potential of LSSA in our further work.

\section*{Acknowledgments}
This work is supported by National Natural Science Foundation (U22B2017) and International Cooperation Foundation of Hubei Province, China (2024EHA032).

\bibliography{main}
\clearpage
\appendix

\section*{Appendix}
\label{sec:appendix}
\input{EMNLP2024/Appendix}

\end{document}

%% file: EMNLP2024/00abstract.tex
Visual-Language Pre-training (VLP) models have achieved significant performance across various downstream tasks. However, they remain vulnerable to adversarial examples. While prior efforts focus on improving the adversarial transferability of multimodal adversarial examples through cross-modal interactions, these approaches suffer from overfitting issues, due to a lack of input diversity by relying excessively on information from adversarial examples in one modality when crafting attacks in another. To address this issue, we draw inspiration from strategies in some adversarial training methods and propose a novel attack called Local Shuffle and Sample-based Attack (LSSA). LSSA randomly shuffles one of the local image blocks, thus expanding the original image-text pairs, generating adversarial images, and sampling around them. Then, it utilizes both the original and sampled images to generate the adversarial texts. Extensive experiments on multiple models and datasets demonstrate that LSSA significantly enhances the transferability of multimodal adversarial examples across diverse VLP models and downstream tasks. Moreover, LSSA outperforms other advanced attacks on Large Vision-Language Models. 


%% file: EMNLP2024/01Introduction.tex
\input{Fig/ALBEFresult}
Visual-Language Pre-training (VLP) models have achieved outstanding performance in various downstream visual-and-language tasks, including image-text retrieval~\cite{ITR, TIR}, visual grounding~\cite{VG}, and image captioning~\cite{IC}.
Despite their success, recent works~\cite{Co-attack,SGA} have revealed that VLP models remain vulnerable to multimodal adversarial examples, which add malicious perturbations to the original image-text pairs.
Moreover, the multimodal adversarial examples crafted for one visual-and-language task are often transferable to other downstream tasks. 
Therefore, it is crucial to explore multimodal adversarial attacks, as they provide valuable insights into the robustness of various VLP models~\cite{ALBEF, TCL, CLIP}.

\input{Fig/CSSA_presentation}
Recent research focuses on attacking VLP models in white-box setting~\cite{Co-attack}, where the attacker has access to the target model's architecture and weights.
However, it is more practical to explore multimodal adversarial attacks in black-box setting~\cite{SGA,Sparse}, where the inner information of target model is not accessible.
As shown in Figure~\ref{fig:ALBEFbar}, while existing works have improved the white-box attack success rate on VLP models through modality interaction, their transferability remains unsatisfied, which is due to overfitting issue caused by limited input diversity, relying excessively on adversarial examples in one modality when crafting attacks in another.

Input transformations have been demonstrated to boost adversarial examples' transferability by enhancing input diversity in unimodal attacks. 
However, recent research~\cite{SGA} reveals that existing unimodal input transformation attacks not only fail to enhance adversarial transferability significantly but also considerably degrade white-box attack performance in multimodal case. 
It is because multimodal retrieval tasks rely more on the spatial information in images and feature alignment~\cite{ITRsurvey}, while unimodal classification tasks depend more on global features.
The data augmentation techniques used in adversarial training also contribute to enhancing input diversity.
However, directly applying unimodal data augmentation to a multimodal attack harms the white-box attack performance due to the fundamental differences between multimodal and unimodal tasks.
On the other hand, the generalization of models can be improved with diverse training data, and the transferability of adversarial examples can be enhanced by leveraging more data information. 
Nonetheless, existing attacks typically consider the adversarial image in generating adversarial text, ignoring the valuable information in the original image.

To address these issues, we draw inspiration from strategies in adversarial training methods~\cite{RLFAT, STAT} and propose a novel attack called Local Shuffle and Sample-based Attack (LSSA), which improves input diversity while preserving spatial information. 
As in Figure~\ref{fig:CSSA_presentation}, LSSA randomly shuffles one of the local image blocks and utilizes the shuffled images and original texts to generate adversarial images.
Then, we sample the neighborhoods around the generated adversarial images and craft adversarial texts using sampled images, original images, and texts. 

We conduct experiments on Flickr30K~\cite{Flickr30k} and MSCOCO~\cite{MSCOCO} to evaluate our LSSA across various downstream tasks.
Experimental results demonstrate that LSSA boosts the attack performance in multimodal learning, outperforming the advanced multimodal attacks in black-box setting.
Additionally, LSSA also surpasses advanced attacks in image captioning and visual grounding.
Especially, to our knowledge, this is the first work to evaluate existing multimodal adversarial transferability performance on Large Vision-Language Models (LVLMs). 
Our contributions are summarized as follows:
\begin{itemize}
    \item We observe that multimodal adversarial attacks rely on input diversity. For images, transformations should maintain spatial information to preserve white-box performance. For texts, using more data can disrupt feature alignment and enhance attack effectiveness.
    \item Based on the observations, we propose a novel attack called Local Shuffle and Sample-based Attack (LSSA), which randomly shuffles one of the local image blocks and samples around adversarial examples to generate adversarial images and texts.
    \item Extensive experiments demonstrate the effectiveness of LSSA, showing a significant improvement in the transferability of adversarial examples across various VLP models and downstream tasks. Moreover, LSSA outperforms advanced attacks on various LVLMs.
\end{itemize}

%% file: Fig/ALBEFresult.tex
\begin{figure}[t]
    \centering
    \includegraphics[width=\columnwidth]{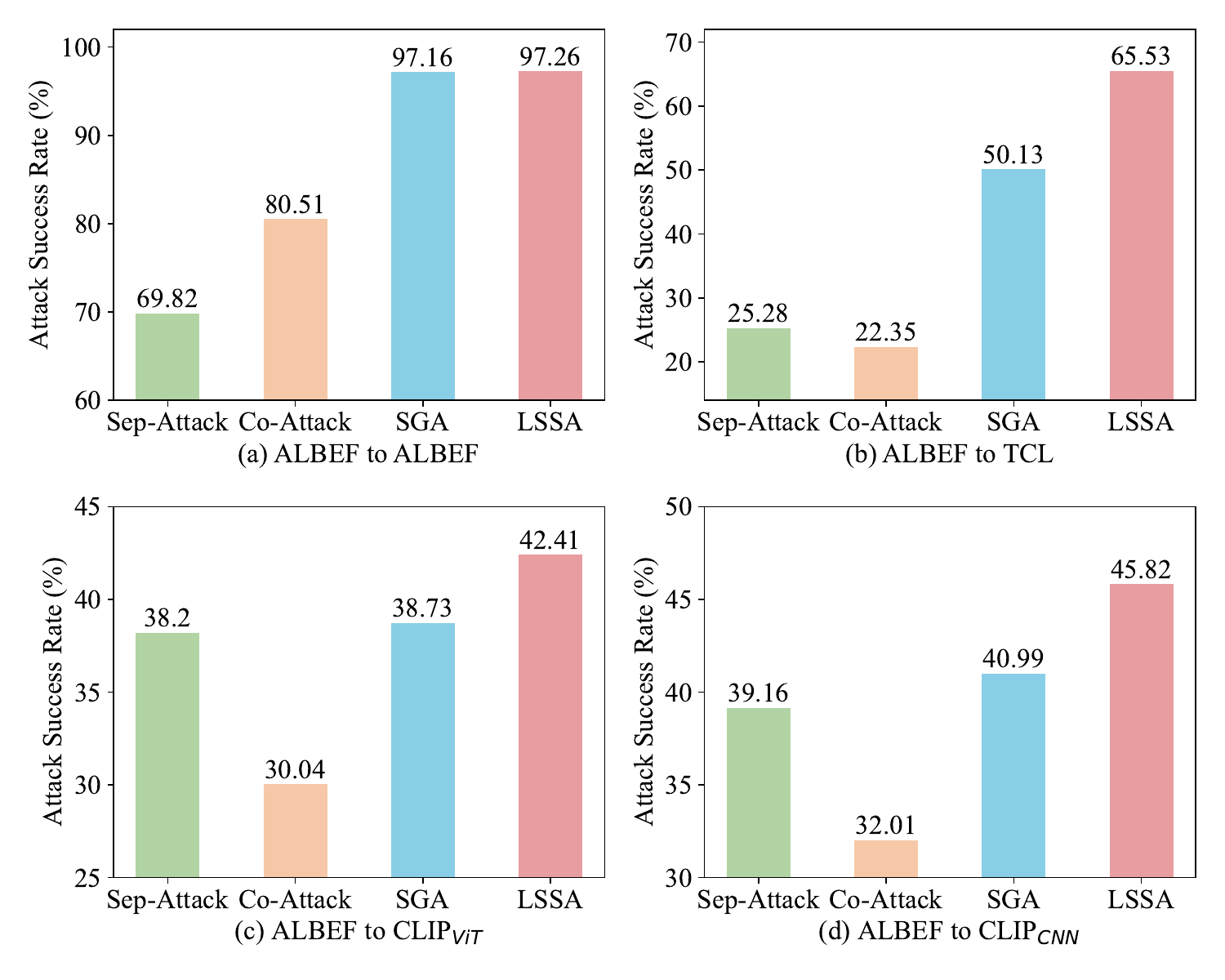}
    \caption{ 
    Comparison of attack success rate (\%) using our LSSA method and existing advanced attacks in image-text retrieval tasks. The multimodal adversarial examples are crafted on the ALBEF model to attack ALBEF, TCL, CLIP\textsubscript{ViT} and CLIP\textsubscript{CNN}, respectively.
    }
    \label{fig:ALBEFbar}
\end{figure}

%% file: Fig/CSSA_presentation.tex
\begin{figure*}[t]
    \centering
    \includegraphics[width=\textwidth]{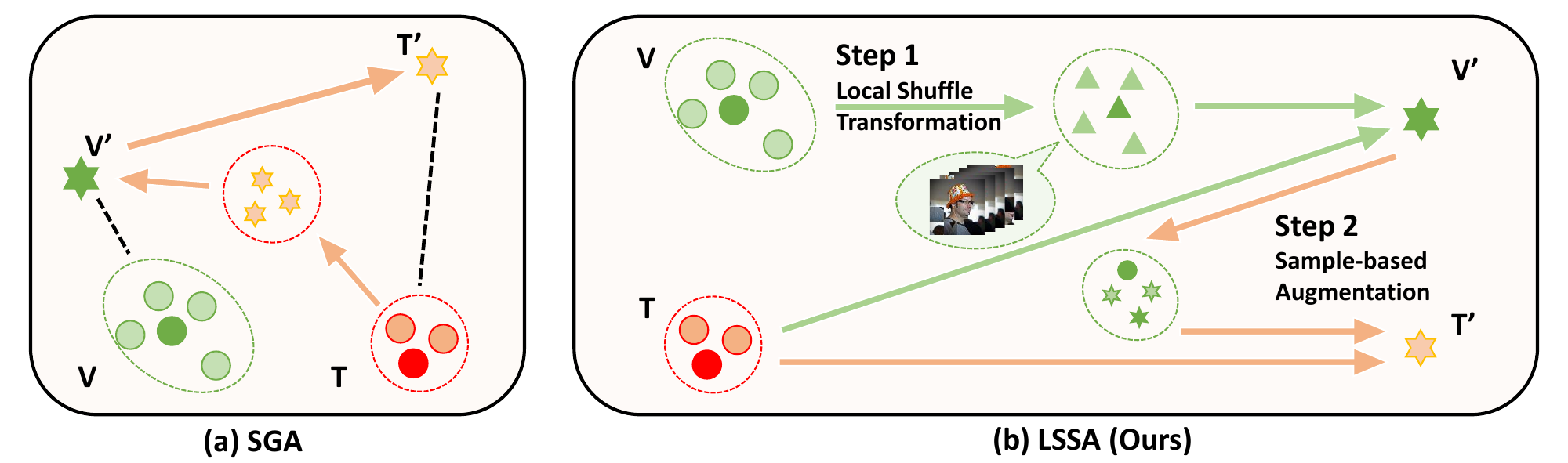}
    \caption{ 
    Comparison of SGA and LSSA. (a) and (b) show the multimodal adversarial examples generation process of SGA and LSSA, respectively. 
    V\textsuperscript{'} and T\textsuperscript{'} represent the corresponding image and text adversarial examples.
    Dashed lines enclose the expanded set.
    Arrows indicate the process or guidance for generating adversarial examples.
    }
    \label{fig:CSSA_presentation}
\end{figure*}

%% file: EMNLP2024/02RW.tex
\subsection{Vision-Language Pre-training Models}
Vision-Language Pre-training (VLP) models boost the performance of various Vision-and-Language tasks~\cite{BLIP} through pre-training on extensive image-text pairs. 
Previous works primarily depend on the pre-trained object detectors to obtain the vision-language representations~\cite{UNITER,Oscar,VLMixer,VinVL}. 
Recently, with the great success of Vision Transformer (ViT)
~\cite{ViT,DBLP:conf/icml/TouvronCDMSJ21,DBLP:conf/iccv/0007CWYSJTFY21} in various tasks, recent works have introduced ViT as an image encoder to convert the input into patches, substituting original computationally expensive object detectors~\cite{DBLP:conf/cvpr/DouXGWWWZZYP0022,BLIP,ALBEF,DBLP:conf/cvpr/WangGZCSQL23,TCL}.

According to the architectures, VLP models can be categorized into fused and aligned models.
Fused VLP models (\textit{e.g.}, ALBEF~\cite{ALBEF}, TCL~\cite{TCL}) initially utilize separate unimodal encoders to obtain visual and text features.
Then, it utilizes a multimodal encoder to output multimodal embeddings by further fusing the embeddings of images and texts. 
In contrast, the aligned VLP model (\textit{e.g.}, CLIP~\cite{CLIP}) extracts visual and text features by separate unimodal encoders and directly aligns their embeddings.
Furthermore, the LVLMs~\cite{BLIP2,MiniGPT-4} also show remarkable performance in numerous multimodal tasks.

\subsection{Downstream Vision-and-Language Tasks}
Given input from one modality, Image-Text Retrieval (ITR)~\cite{ITR,TIR} is a retrieval task where the goal is to retrieve the most relevant instances from a gallery database in the other modality. 

For fused VLP models~\cite{ALBEF,TCL,BLIP}, the similarity scores are calculated for all image-text pairs to retrieve top-k candidates, which are then processed by a multimodal encoder to compute final image-text matching scores for ranking.
For aligned VLP model~\cite{CLIP}, the final ranking can be directly determined by the image and text embedding similarity.

Image Captioning (IC)~\cite{IC} generates suitable and logical textual descriptions for the visual images. 
The evaluation metrics for image captioning models, including BLEU~\cite{BLEU}, METEOR~\cite{METEOR}, ROUGE~\cite{ROUGE}, CIDEr~\cite{CIDEr}, and SPICE~\cite{SPICE}, compare the generated text and the reference text in terms of quality and relevance. 
Visual Grounding (VG)~\cite{VG} is a localization task that identifies the corresponding visual region in images based on the textual descriptions.

\subsection{Transferability of Adversarial Examples}
Existing adversarial attacks can be categorized into white-box and black-box attacks.
In white-box setting, the attacker has full access to the information of the target model, while such access is not available in black-box settings.
In computer vision, numerous attacks have been proposed to craft adversarial examples by utilizing gradient information or input transformations, such as FGSM~\cite{fgsm}, MIM~\cite{MIM}, and TIM~\cite{TIM}.
In natural language processing, existing attacks, such as BERT-Attack~\cite{bertattack} and FGPM~\cite{FGPM}, modify characters and words in the inputs.

In the multimodal vision-language field, Co-attack~\cite{Co-attack} is the first and typical white-box attack for popular VLP models on downstream tasks.
SGA~\cite{SGA} diversifies image-text pairs and extends the embedding distance of image-text pairs to improve the transferability of multimodal adversarial examples in black-box setting.
However, SGA suffers from the overfitting issue due to a lack of input diversity by relying excessively on information from adversarial examples in one modality when crafting attacks in another. 
To further improve the adversarial transferability, we enrich the diversity of adversarial examples for image-text pairs and consider both original and sampled image-text pairs, enabling the transferability of adversarial examples.

%% file: EMNLP2024/03Method.tex
This section first introduces our motivations and observations.
Then, we propose a novel multimodal adversarial attack called Local Shuffle and Sample-based Attack, providing a detailed description.

\subsection{Motivations and Observations}
\label{section:3.1}
\input{Table/shuffle_motivation}
Existing unimodal adversarial attacks, such as DIM~\cite{DIM}, SIM~\cite{NI}, and multimodal attack SGA~\cite{SGA} use input transformation to effectively increase the diversity of inputs, thereby improving the performance of the attacks. 
However, SGA shows that existing unimodal adversarial attacks based on input transformation significantly harm the white-box attack performance.
On the other hand, we observe that data augmentation in adversarial training can also effectively increase input diversity. 
Therefore, inspired by the adversarial training method of RLFAT~\cite{RLFAT}, we conduct experiments to explore the impact of data augmentation on the transferability of adversarial examples.

Specifically, we combine the Global Shuffle of RLFAT with SGA. 
Following their settings, we also divide the original images into blocks and randomly shuffle them to generate adversarial examples using the gradients of shuffled images. 
Detailed information is described in Appendix~\ref{Visualization of Shuffle Images} and Appendix~\ref{Visualization of Multimodal Adversarial Examples}.
As shown in Table~\ref{tab:shuffle_motivation}, Global Shuffle improves the adversarial transferability, yet the performance of white-box attacks significantly decays.
It is because multimodal retrieval tasks rely more on the spatial information in images and feature alignment while Global Shuffle excessively disrupts them in the original images.
Therefore, a suitable input transformation needs to increase diversity while preserving the original spatial information.
To address this issue, we block the original image and randomly shuffle one of the local image blocks.
As shown in Table~\ref{tab:shuffle_motivation}, Local Shuffle transformation significantly enhances the black-box attack while maintaining white-box performance, boosting the diversity of original image-text pairs \((v,t)\).

\input{Fig/sample_motivation}
In the text modality, previous attacks only consider either the original image or the adversarial image in generating adversarial text, without considering both information. 
Similar to improving the generalization of models with more training data, the transferability of adversarial examples can be enhanced by utilizing more data information~\cite{NI}.
Therefore, it is natural to consider the information from multiple samples to improve the adversarial transferability. 
Inspired by another adversarial training method of STAT~\cite{STAT}, which utilizes both the original and the neighbors of the adversarial examples to enhance the model robustness, we conduct experiments to investigate the multimodal adversarial examples generated by different images and text information. 
As illustrated in Figure~\ref{fig:sample_motivation}, the experimental results indicate that directly using the original text outperforms using adversarial text for generating transferable adversarial images. 
Additionally, combining the original image with the sampled points around the adversarial image shows better transferability compared to individually using the adversarial image.

\subsection{Local Shuffle and Sample-based Attack}
\textbf{Local Shuffle Transformation.}
Based on the analysis in Section~\ref{section:3.1}, in each iteration, we randomly shuffle one of the local image blocks of the current adversarial images \((v^{adv}_i, t)\), and repeat to get $N$ samples.
By utilizing the above input transformation method, we can obtain an expanded dataset \((V, T)={\{(v^{'}_1,t),(v^{'}_2,t),...,(v^{'}_N,t)\}}\).
Then, the expanded dataset \((V,T)\) is used to generate the adversarial images \((v^{adv}_{i+1}, t)\), which can be formulated as follows:
\begin{gather}
\begin{split}
    \small
    g_{i+1}&=\mu g_i+\frac{1}{N}\sum^{N}_{j=1}\frac{\nabla_{v}J(f_I(v^{'}_j),f_T(t))}{\|\nabla_{v}J(f_I(v^{'}_j),f_T(t)\|}, \\
    v^{adv}_{i+1}&=Clip_{\epsilon_v}(v^{adv}_i+\alpha \cdot \textit{sign}(g_{i+1})),
    \end{split}
\end{gather}
where \(J(\cdot,\cdot)\) is the loss function, \(g_i\) is the accumulated gradient, \(\epsilon_v\) is the perturbation boundary of the image, \(\alpha\) is the step size and \(\mu\) is the decay factor.
\(f_I(\cdot)\) and \(f_T(\cdot)\) are visual and text encoder.

\textbf{Sample-based Augmentation.}
Based on the 
analysis in Section~\ref{section:3.1}, it is evident that the original and neighbors of adversarial images can effectively enhance the transferability of multimodal adversarial examples.
It ensures the adversarial text has significant differences not only from the original image features but also from the adversarial image features. 
The adversarial text generation process can be formalized as follows:
\begin{equation}
\begin{split}
    t^{adv}&=\underset{t^{'}\in B[t,\epsilon_t]}{\arg\max} \quad \lambda \cdot J(f_I(v),f_T(t^{\prime}))\\
    & +(1-\lambda)\cdot \frac{1}{M}\sum_{i=1}^{M}J(f_I(v_i),f_T(t^{\prime})),
\end{split}
\end{equation}
where \(J(\cdot,\cdot)\) is the loss function, \(M\) is the sampled number, \(v\) is the original image, and \(v_i\) is the neighbours of adversarial image.
The overall algorithm of LSSA is summarized in Appendix~\ref{alg_CSSA}.

\textbf{Difference of LSSA.} Since our LSSA attack draws inspiration from the strategies of adversarial training methods RLFAT and STAT, we highlight the differences as follows:

\textbf{a) } The goal of adversarial training is to boost model generalization, while our LSSA aims to generate more transferable adversarial examples with input diversity.
\textbf{b) } The local shuffle in our LSSA is designed for multimodal retrieval tasks, whereas the global shuffle in RLFAT is proposed for unimodal classification tasks.
\textbf{c) } The sample-based attack enhances performance by randomly sampling around the adversarial images, whereas STAT simultaneously crafts different adversarial images to improve model robustness.

With different goals, tasks, and methods, our LSSA is a new and novel input transformation based multimodal adversarial attack to improve the transferability of attack performance. 

%% file: Table/shuffle_motivation.tex
\begin{table*}[t]
\caption{The attack success rate (\%) of multimodal adversarial examples against different VLP models. The source column represents the VLP models used for crafting multimodal adversarial examples on the Flickr30K dataset. * indicates white-box attacks. The higher attack success rate indicates the better performance.
}
\centering
\resizebox{\textwidth}{!}{
\begin{threeparttable}
\begin{tabular}{l|l|cc|cc|cc|cc}
\toprule
\multirow{2}{*}{Source} & \multirow{2}{*}{Attack} &
  \multicolumn{2}{c}{ALBEF} &
  \multicolumn{2}{c}{TCL} &
  \multicolumn{2}{c}{CLIP\textsubscript{ViT}} &
  \multicolumn{2}{c}{CLIP\textsubscript{CNN}} 
  \\
    \cline{3-10}
  & &TR R@1& IR R@1&TR R@1& IR R@1&TR R@1& IR R@1&TR R@1& IR R@1 \\
  \midrule
  \multirow{3}{*}{ALBEF}
  &SGA&\textbf{97.24}\textsuperscript{*}&\textbf{97.08}\textsuperscript{*}&44.68&55.57&33.25&44.20&35.38&46.59\\
  &Global Shuffle&91.97\textsuperscript{*}&91.93\textsuperscript{*}&48.26&58.02&34.87&46.79&38.85&48.95\\
&Local Shuffle&96.45\textsuperscript{*}&96.40\textsuperscript{*}&\textbf{56.16}&\textbf{64.38}&\textbf{35.95}&\textbf{48.16}&\textbf{40.38}&\textbf{50.23}\\
\bottomrule
\end{tabular}
\end{threeparttable}
}
\label{tab:shuffle_motivation}
\end{table*}

%% file: Fig/sample_motivation.tex
\begin{figure}[t]
    \centering
    \includegraphics[width=\columnwidth]{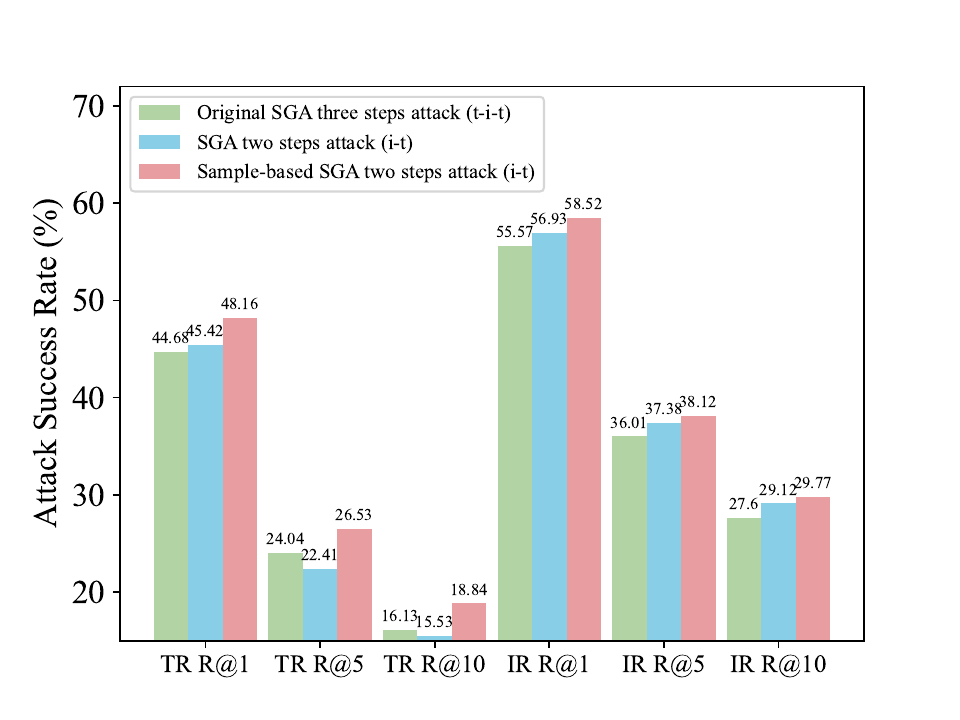}
    \caption{The attack success rate of multimodal adversarial examples against TCL model, which are crafted on ALBEF model. 
    We explore three methods: \textbf{a) Original SGA three steps attack (t-i-t)}, \textbf{b) SGA two steps attack (i-t)}: generating adversarial images using original text and then generating adversarial text using the adversarial images, \textbf{c) Sample-based SGA two steps attack (i-t)}: generating adversarial images using original text and then using the original and neighbours of the adversarial image to generate the adversarial images. 
    }
    \label{fig:sample_motivation}
\end{figure}

%% file: EMNLP2024/04Experiments.tex
In this section, we conduct experiments on two benchmark datasets using popular VLP models and LVLMs. First, we describe the experimental settings in Section~\ref{sec:4.1}. Next, we evaluate the performance of various advanced multimodal adversarial attacks in Section~\ref{sec:4.2}. We then analyze the cross-task transferability between different V+L tasks in Section~\ref{sec:4.3} and present the results of further study in Section~\ref{sec:4.4}. Finally, we evaluate the LSSA method on various LVLMs to verify its generalization in Section~\ref{sec4.5}.

\subsection{Experimental Settings}
\textbf{Datasets and VLP Models.}
\label{sec:4.1}
Following the setting on SGA~\cite{SGA}, we conduct experiments on two benchmark datasets, namely Flickr30K~\cite{Flickr30k} and MSCOCO~\cite{MSCOCO}.
Flickr30K and MSCOCO dataset contains 31,783 and 123,287 images, respectively, with five corresponding captions.
We evaluate two typical architectures of VLP models: the fused VLP models and aligned models.
For the fused VLP models, we choose ALBEF~\cite{ALBEF} and TCL~\cite{TCL}, which are pre-trained by different pre-trained tasks.
For the aligned VLP models, we choose the CLIP model with different visual encoders.
More details are provided in Appendix~\ref{Experimental Setting}.

\textbf{Adversarial Attack Settings.}
We follow the multimodal adversarial attack setting of SGA. 
Moreover, we extend the dataset by random shuffling one of the local image blocks, where the size of the original images is \(h \times w\), and the size of the local image block is \(h/2 \times w/2\), which is blocked to four \(h/4 \times w/4\) subblocks.  
We randomly shuffle \(N=20\) times for each image.
For the sampled image number, we choose \(M=20\) to generate text adversarial examples.
More details are provided in Appendix~\ref{Experimental Setting}.

\input{Table/CSSA_main}
\subsection{Experiments Results}
\label{sec:4.2}
We conduct extensive experiments to evaluate our LSSA method on two widely adopted architectures of models: fused VLP and aligned VLP models.

As shown in Table~\ref{tab:CSSA_main}, our proposed LSSA outperforms existing advanced multimodal attacks in white-box and black-box settings.
When the type of source model is the same as the target model, LSSA has significantly improved adversarial transferability.
Specifically, LSSA surpasses SGA by 14.75\% and 10.05\% in terms of attack success rate on TR and IR tasks, respectively, when transferring the multimodel adversarial examples from ALBEF to TCL.
We observe the same phenomenon that the multimodal adversarial examples have better adversarial transferability from CLIP\textsubscript{ViT} to CLIP\textsubscript{CNN}, which outperforms SGA by 7.67\% and 7.89\% on TR and IR tasks, respectively.
Moreover, we also evaluate the performance of LSSA on another benchmark dataset MSCOCO, which is more challenging.
Our LSSA still exceeds the adversarial transferability of baseline methods with a clear margin. 
The details of experiments on the MSCOCO dataset are provided in Appendix~\ref{MSCOCO_main}.

To explore the effectiveness of LSSA on the different architecture of VLP models, we craft multimodal adversarial examples from ALBEF (CLIP) to CLIP (ALBEF) model.
As shown in Table~\ref{tab:CSSA_main}, our proposed LSSA consistently improves the transferability of adversarial attacks on both ViT-based CLIP and ViT-based CLIP, generated on ALBEF model, surpassing the performance of other advanced attacks.
Specifically, LSSA significantly enhances the adversarial transferability, attaining an improvement of 3.07\%, 4.69\% and 5.59\%, 4.18\% on CLIP\textsubscript{ViT} and CLIP\textsubscript{CNN}.

\subsection{Cross-Task Transferability}
\label{sec:4.3}
We also conduct extensive experiments to evaluate the effectiveness of our LSSA attack on Image Captioning (IC) and Visual Grounding (VG).

In the IC task, we craft adversarial images in the ALBEF model and then attack the BLIP~\cite{BLIP} model on the MSCOCO dataset.
As shown in Table~\ref{tab:CSSA_IC}, LSSA enhances the adversarial transferability compared with SGA, gaining improvements of 2.87\%, 4.93\%, 2.49\%, 6.29\% and 6.54\%, respectively.
\input{Table/CSSA_IC}
In the VG task, we generate adversarial image-text pairs on the RefCOCO+ dataset.
The source and target models are ALBEF models pre-trained on the ITR and VG tasks, respectively.
As shown in Table~\ref{tab:CSSA_VG}, LSSA still performs better than existing attacks, significantly improving 8.58\%, 12.42\%, and 6.18\%, respectively.
It might be because the adversarial perturbations crafted by LSSA contain more spacial adversarial information compared with other modal interaction attacks. 
The results demonstrate the effectiveness of our LSSA on cross-task transferability. 
\input{Table/CSSA_VG}

\subsection{Further Study}
\label{sec:4.4}
We conduct various experiments on image-text retrieval to explore the impact of hyper-parameters in the LSSA method, including the number of local shuffles, the positions of local shuffles, and the weight of the loss. 
Detailed information of the ablation study is provided in Appendix~\ref{Abaltion study}.
Specifically, we use ALBEF as the white-box model to craft multimodal adversarial examples, while other models are attacked as black-box models.

\textbf{Number of local shuffles \(N\).}
\input{Fig/shuffle_num}
We conduct experiments to explore the effectiveness of local shuffle number \(N\). 
Specifically, when the number of local shuffles \(N=0\), it degenerates into a sample-based attack.
As shown in Figure~\ref{fig:shuffle_num}, local shuffle can effectively enhance the transferability of multimodal adversarial examples, and the enhancement improves with the increase of the number of shuffled samples, reaching its maximum at approximately \(N=20\). 
Specifically, compared to no local shuffle, shuffling gains an improvement of 5.59\%, 1.96\%, 1.15\% on IR task and 4.69\%, 1.33\%, 0.72\% on TR task, crafted multimodal adversarial examples in TCL, CLIP\textsubscript{ViT} and CLIP\textsubscript{CNN}. 
To achieve optimal performance, we choose \(N=20\).

\textbf{Positions of local shuffle.}
\input{Fig/shuffle_idx}
We conduct experiments to investigate the impact of the shuffle position on the diversity of image-text pairs and the transferability of multimodal adversarial examples.
Specifically, we only shuffle the left top, right top, left bottom, and right bottom of the image. 
Moreover, we compare the performance of the above strategies with a random shuffle.
As shown in Figure~\ref{fig:shuffle_idx}, shuffling different positions results in variations in the generated multimodal adversarial examples.
Moreover, the impact of local shuffling differs across different VLP models, possibly due to variations in the model's attention to different regions of the image-text pairs. 
To mitigate this influence and improve the attack performance, we choose the random shuffle strategy.

\textbf{Weight of the loss \(\lambda\).}
\input{Fig/loss_weight}
The weight of loss \(\lambda\) controls the trade-off between the loss of original image-pairs and sampled image-pairs.
To investigate the impact of weight loss \(\lambda\), we evaluate the effectiveness with different \(\lambda\).
Specifically, we let \(\lambda\) vary from 0 to 1. 
When \(\lambda = 0\), the loss disregards the loss of the original image-text pairs, while when \(\lambda=1\), the loss disregards the information from the sampled image-text pairs.
As shown in Figure~\ref{fig:loss_weight}, the performance of LSSA increases and reaches the peak when \(\lambda=0.5\), which indicates that excessive reliance on either the information from the original image-text pairs or the sampled image-text pairs can lead to a decrease in the attack success rate.
Therefore, we select an intermediate value of \(\lambda=0.5\) to obtain the best performance.

\subsection{Attack Performance on LVLMs}
\input{Table/CSSA_LVLMs}
\label{sec4.5}
We first evaluate the attack performance of existing attacks and our LSSA on LVLMs, which have achieved remarkable success on various tasks.
Specifically, we conduct attacks on BLIP-2~\cite{BLIP2}, VisualGLM~\cite{visualGLM}, MiniGPT4~\cite{MiniGPT-4} and PandaGPT~\cite{PandaGPT} in black-box setting. 
Due to the difference between VLP and LVLM tasks, we utilize dialogues to compare the images and texts.
For each image, we use a language template ("Question: Does the picture depict that \textit{[text]}? Answer:") to ask LVLMs to obtain judgments.
As shown in Table~\ref{tab:CSSA_LVLMS}, LSSA still surpasses baselines, gaining an improvement of 1.19\%, 0.80\%, 2.02\%, and 1.20\%. 
It indicates the effectiveness and robustness of LSSA compared with other advanced attacks.

%% file: Table/CSSA_main.tex
\begin{table*}[t]
\caption{The attack success rate (\%) of multimodal adversarial examples against different VLP models compared with state-of-the-art methods on image-text retrieval task. The source column represents the VLP models used for crafting multimodal adversarial examples on the Flickr30K dataset. * indicates white-box attacks. 
}
\centering
\resizebox{\textwidth}{!}{
\begin{threeparttable}
\begin{tabular}{l|l|cc|cc|cc|cc}
\toprule
\multirow{2}{*}{Source} & \multirow{2}{*}{Attack} &
  \multicolumn{2}{c}{ALBEF} &
  \multicolumn{2}{c}{TCL} &
  \multicolumn{2}{c}{CLIP\textsubscript{ViT}} &
  \multicolumn{2}{c}{CLIP\textsubscript{CNN}} 
  \\
    \cline{3-10}
  & &TR R@1& IR R@1&TR R@1& IR R@1&TR R@1& IR R@1&TR R@1& IR R@1 \\
  \midrule
  \multirow{6}{*}{ALBEF}
&PGD&52.45\textsuperscript{*}&58.65\textsuperscript{*}&~~3.06&~~6.79&~~8.96&13.21&10.34&14.65\\
&BERT-Attack&11.57\textsuperscript{*}&27.46\textsuperscript{*}&12.64&28.07&29.33&43.17&32.69&46.11\\
&Sep-Attack&65.69\textsuperscript{*}&73.95\textsuperscript{*}&17.60&32.95&31.17&45.23&32.82&45.49\\
&Co-Attack&77.16\textsuperscript{*}&83.86\textsuperscript{*}&15.21&29.49&23.60&36.48&25.12&38.89\\
&SGA&97.24\textsuperscript{*}&97.08\textsuperscript{*}&44.68&55.57&33.25&44.20&35.38&46.59\\
&LSSA (Ours)&\textbf{97.29\textsuperscript{*}}&\textbf{97.22\textsuperscript{*}}&\textbf{59.43}&\textbf{65.62}&\textbf{36.32}&\textbf{48.49}&\textbf{40.87}&\textbf{50.77}\\
  \midrule
  \multirow{6}{*}{TCL}
&PGD&~~6.15&10.78&77.87\textsuperscript{*}&79.48\textsuperscript{*}&~~7.48&13.72&10.34&15.33\\
&BERT-Attack&11.89&26.82&14.54\textsuperscript{*}&29.17\textsuperscript{*}&29.69&44.49&33.46&46.07\\
&Sep-Attack&20.13&36.48&84.72\textsuperscript{*}&86.07\textsuperscript{*}&31.29&44.65&33.33&45.80\\
&Co-Attack&23.15&40.04&77.94\textsuperscript{*}&85.59\textsuperscript{*}&27.85&41.19&30.74&44.11\\
&SGA&49.32&59.92&98.52\textsuperscript{*}&98.79\textsuperscript{*}&33.87&45.07&38.44&47.92\\
&LSSA (Ours)&\textbf{59.85}&\textbf{67.59}&\textbf{98.63\textsuperscript{*}}&\textbf{98.80\textsuperscript{*}}&\textbf{35.46}&\textbf{48.39}&\textbf{41.12}&\textbf{51.05}\\
  \midrule
  \multirow{6}{*}{CLIP\textsubscript{ViT}}
&PGD&~~2.50&~~4.93&~~4.85&~~8.17&70.92\textsuperscript{*}&78.61\textsuperscript{*}&~~5.36&~~8.44\\
&BERT-Attack&~~9.59&22.64&11.80&25.07&28.34\textsuperscript{*}&39.08\textsuperscript{*}&30.40&37.43\\
&Sep-Attack&~~9.59&53.25&11.38&25.60&79.75\textsuperscript{*}&86.79\textsuperscript{*}&30.78&39.76\\
&Co-Attack&10.57&24.33&11.94&26.69&93.25\textsuperscript{*}&95.86\textsuperscript{*}&32.52&41.82\\
&SGA&13.56&27.01&14.54&30.07&98.80\textsuperscript{*}&98.94\textsuperscript{*}&40.61&47.55\\
&LSSA (Ours)&\textbf{16.89}&\textbf{32.11}&\textbf{17.18}&\textbf{33.07}&\textbf{98.85\textsuperscript{*}}&\textbf{98.97\textsuperscript{*}}&\textbf{48.28}&\textbf{55.44}\\
  \midrule
  \multirow{6}{*}{CLIP\textsubscript{CNN}}
&PGD&~~2.09&~~4.82&~~4.00&~~7.81&~~1.10&~~6.60&86.46\textsuperscript{*}&92.25\textsuperscript{*}\\
&BERT-Attack&~~8.86&23.27&12.33&25.48&27.12&37.44&30.40\textsuperscript{*}&40.10\textsuperscript{*}\\
&Sep-Attack&~~8.55&23.41&12.64&26.12&28.34&39.43&91.44\textsuperscript{*}&95.44\textsuperscript{*}\\
&Co-Attack&~~8.79&23.79&13.10&26.07&28.79&40.03&94.76\textsuperscript{*}&96.89\textsuperscript{*}\\
&SGA&10.74&25.02&14.54&27.26&30.80&41.82&99.23\textsuperscript{*}&99.42\textsuperscript{*}\\
&LSSA (Ours)&\textbf{14.08}&\textbf{26.89}&\textbf{15.28}&\textbf{30.12}&\textbf{37.67}&\textbf{46.91}&\textbf{99.74\textsuperscript{*}}&\textbf{99.97\textsuperscript{*}}\\
\bottomrule
\end{tabular}
\end{threeparttable}
}
\label{tab:CSSA_main}
\end{table*}

%% file: Table/CSSA_IC.tex
\begin{table}[t]
\caption{\textbf{Cross-Task Transerability: ITR \(\rightarrow\) IC.} The adversarial image-text pairs are generated from Image-Text Retrieval (ITR) task to attack Image Caption (IC) task on MSCOCO dataset. The source and target model are ALBEF and BLIP, respectively. The Baseline is the performance of IC on the original data, where the lower value indicates better cross-task transerability of adversarial examples.
}
\centering
\resizebox{\columnwidth}{!}{
\begin{threeparttable}
\begin{tabular}{l|ccccc}
\toprule
Attack & B@4 & METEOR & ROUGE\_L & CIDEr & SPICE \\
\midrule
Baseline & 39.7 & 31.0 & 60.0 & 133.3 & 23.8\\
Co-Attack & 37.4 & 29.8 & 58.4 & 125.5 & 22.8\\
SGA & 34.8 & 28.4 & 56.3 & 116.0 & 21.4 \\
LSSA (Ours) & \textbf{33.8} & \textbf{27.0}&\textbf{54.9} &\textbf{108.7} & \textbf{20.0}\\
\bottomrule
\end{tabular}
\end{threeparttable}
}
\label{tab:CSSA_IC}
\end{table}

%% file: Table/CSSA_VG.tex
\begin{table}[t]
\caption{\textbf{Cross-Task Transerability: ITR \(\rightarrow\) VG.} The adversarial image-text pairs are generated from the Image-Text Retrieval (ITR) task to attack the Visual Grounding (VG) task in the ALBEF model on the RefCOCO+ dataset. The Baseline is the performance of VG on the original data, where the lower value indicates better cross-task transferability of adversarial examples.
}
\centering
\resizebox{0.35\textwidth}{!}{
\begin{threeparttable}
\begin{tabular}{l|ccc}
\toprule
Attack & Val & TestA & TestB\\
\midrule
Baseline & 58.5 & 65.9 & 46.3 \\
Co-Attack &54.3&61.8&43.8 \\
SGA & 53.6&61.2&43.7 \\
LSSA (Ours) & \textbf{49.0} & \textbf{53.6}&\textbf{41.0}\\
\bottomrule
\end{tabular}
\end{threeparttable}
}
\label{tab:CSSA_VG}
\end{table}

%% file: Fig/shuffle_num.tex
\begin{figure*}[t]
    \centering
    \subfigure[ALBEF to TCL]{
    \includegraphics[width=0.32\textwidth]{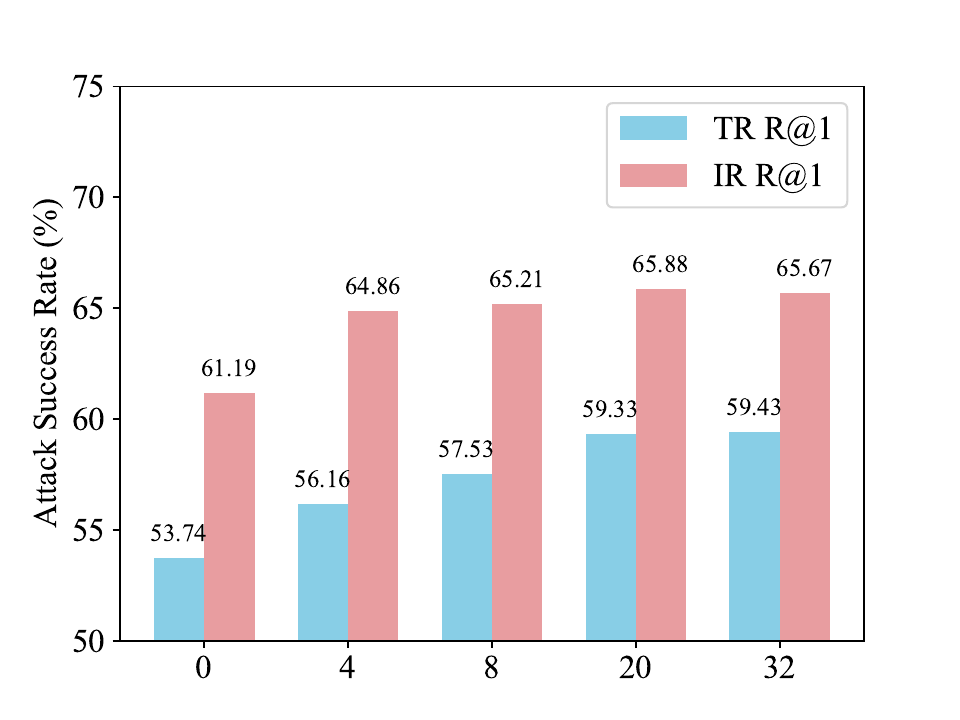}}
    \subfigure[ALBEF to CLIP\textsubscript{ViT}]{
    \includegraphics[width=0.32\textwidth]{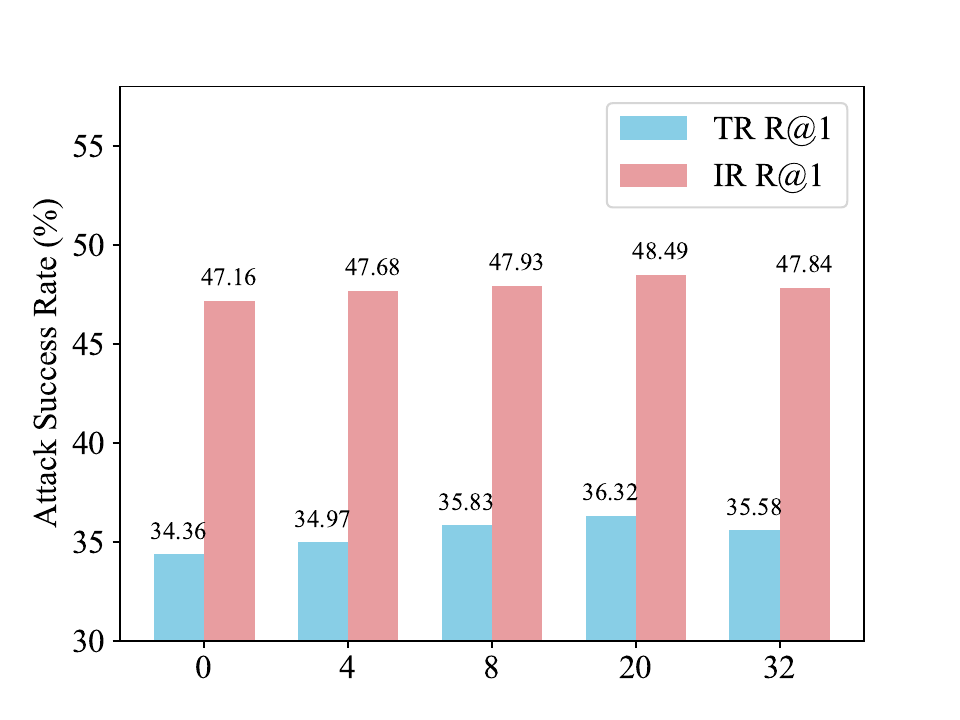}}
    \subfigure[ALBEF to CLIP\textsubscript{CNN}]
    {\includegraphics[width=0.32\textwidth]{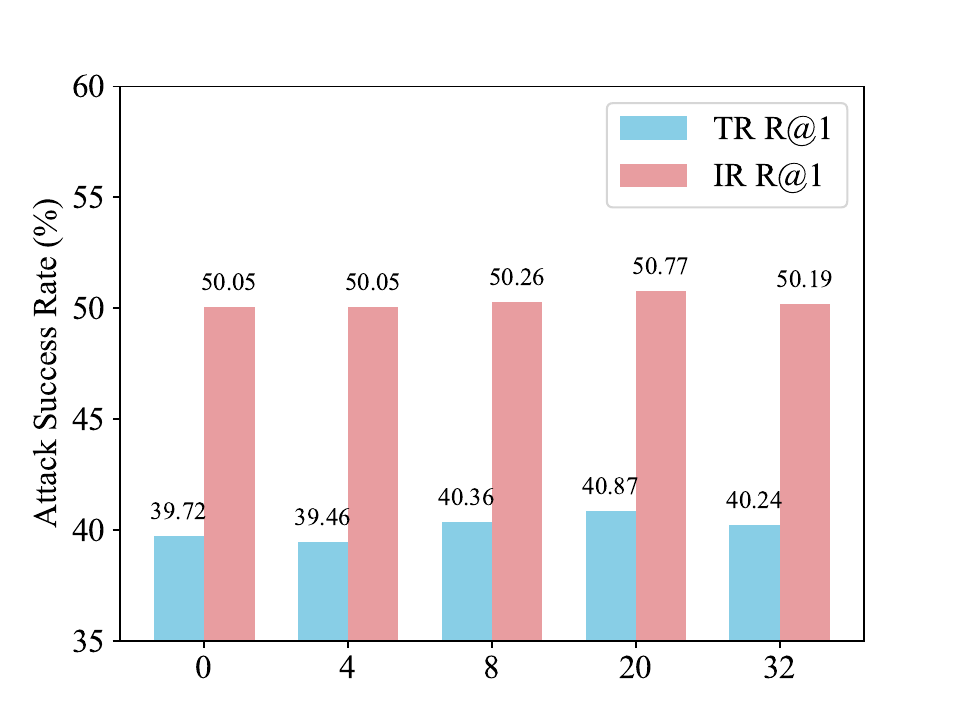}}
    \caption{The attack success rate (\%) of LSSA with different shuffle number \(N\) on the Flickr30K dataset. The source model is ALBEF, and the target model is other VLP models.
    }
    \label{fig:shuffle_num}
\end{figure*}

%% file: Fig/shuffle_idx.tex
\begin{figure*}[t]
    \centering
    \subfigure[ALBEF to TCL]{
    \includegraphics[width=0.32\textwidth]{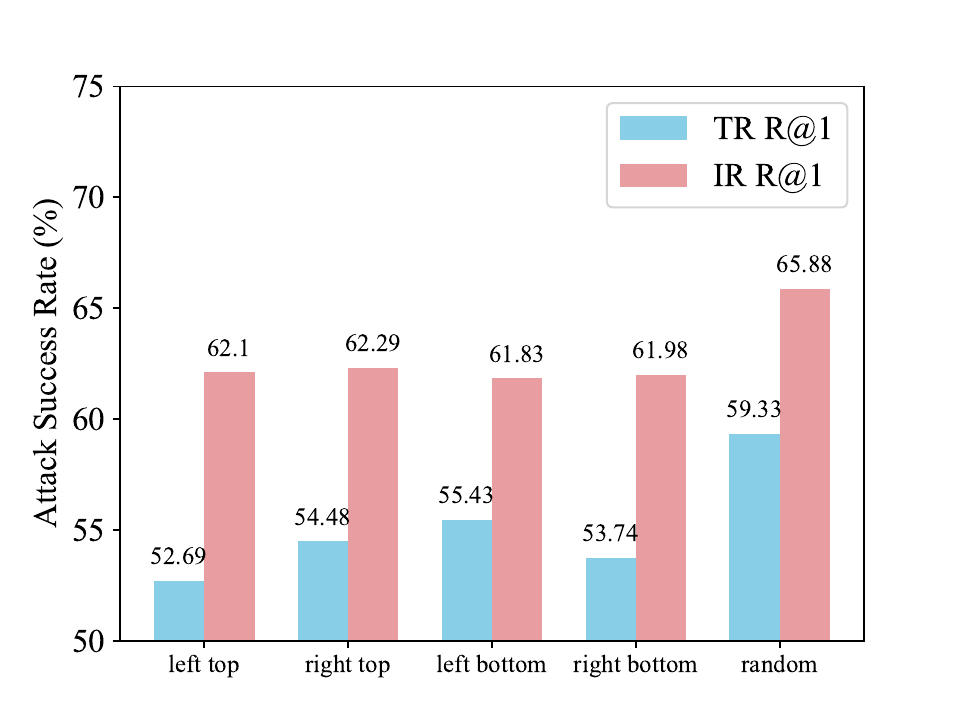}}
    \subfigure[ALBEF to CLIP\textsubscript{ViT}]{
    \includegraphics[width=0.32\textwidth]{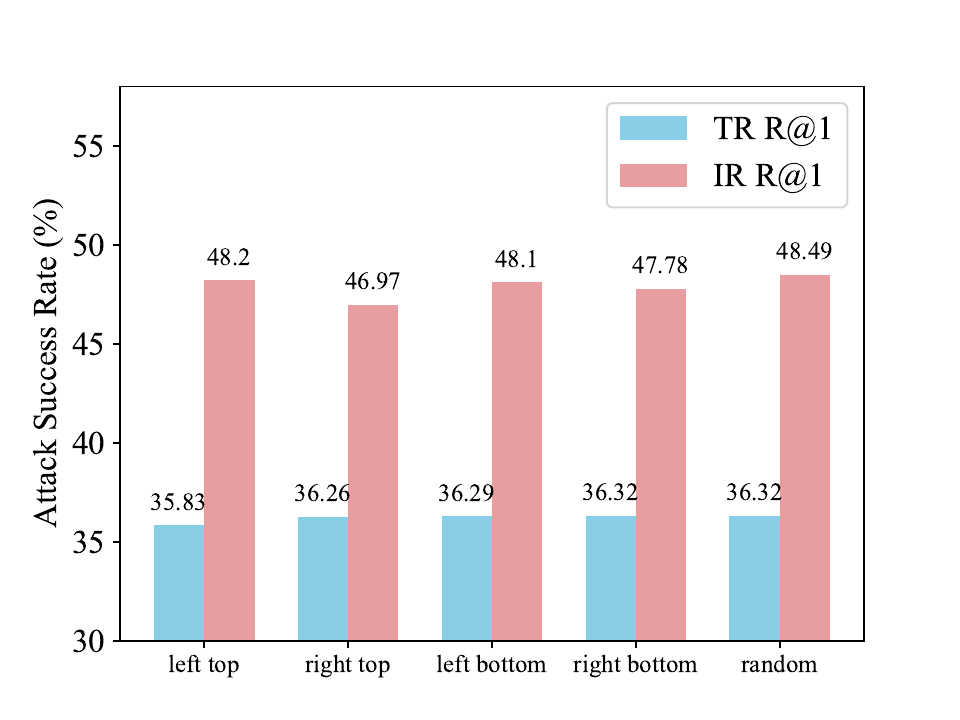}}
    \subfigure[ALBEF to CLIP\textsubscript{CNN}]
    {\includegraphics[width=0.32\textwidth]{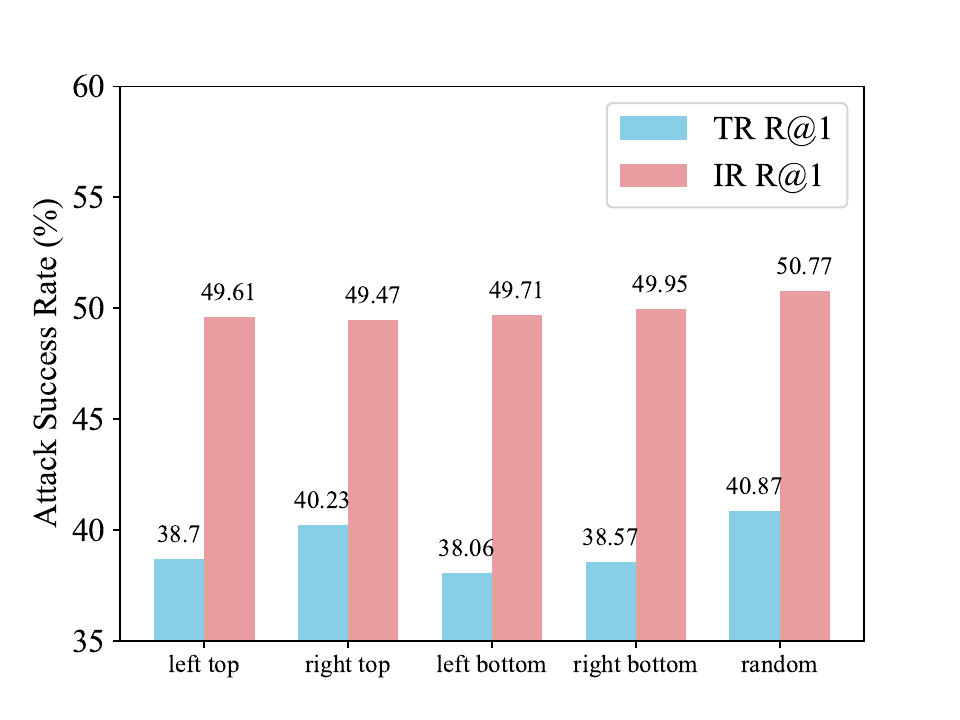}}
    \caption{The attack success rate (\%) of LSSA with different shuffle position on the Flickr30K dataset. The source model is ALBEF, and the target model is other VLP models.
    }
    \label{fig:shuffle_idx}
\end{figure*}

%% file: Fig/loss_weight.tex
\begin{figure}[t]
    \centering
    \includegraphics[width=\columnwidth]{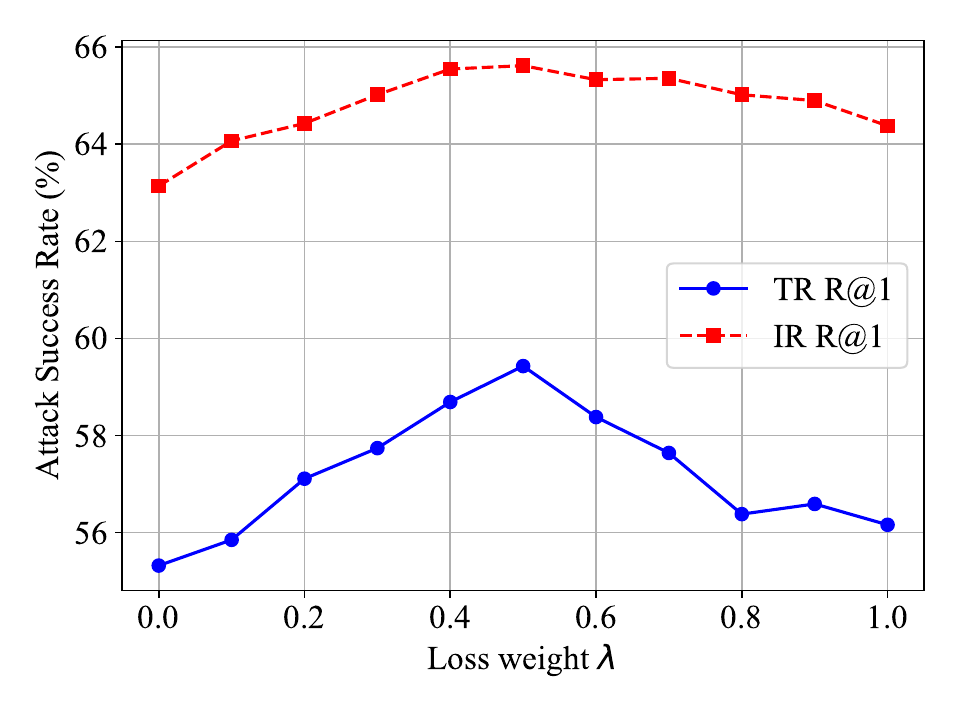}
    \caption{ 
    The attack success rate (\%) of CSSA with different loss weights \(\lambda\) on the Flickr30K dataset. The source and target model are ALBEF and TCL model.
    }
    \label{fig:loss_weight}
\end{figure}

%% file: Table/CSSA_LVLMs.tex
\begin{table}[t]
\caption{The attack success rate (\%) of LSSA for black-box attacks on LVLMs using the Flickr30K dataset, where the surrogate model is the ALBEF model.
}
\centering
\resizebox{\columnwidth}{!}{
\begin{threeparttable}
\begin{tabular}{l|c|c|c|c}
\toprule
Attack/Model & BLIP2 & VisualGLM & MiniGPT4 & PandaGPT \\
\midrule
Baseline & ~~0.84 & ~~0.68 & ~~0.92 & ~~0.86 \\
Co-Attack &19.82& 10.38 & 25.80 &19.06\\
SGA &20.45&10.72&25.54&19.52
\\
LSSA (Ours) &\textbf{21.64}&\textbf{11.52}&\textbf{27.56}&\textbf{20.72}\\
\bottomrule
\end{tabular}
\end{threeparttable}
}
\label{tab:CSSA_LVLMS}
\end{table}

%% file: EMNLP2024/05Conclusion.tex
In this work, we first investigate the adversarial transferability of multimodal attacks.
We observe that existing attacks suffer from the overfitting issue due to a lack of input diversity by relying excessively on information from adversarial examples in one modality when crafting attacks in another.
To address this issue, we propose a novel multimodal adversarial attack called Local Shuffle and Sample-based Attack (LSSA), which utilizes local shuffle to enrich image-text pairs and generate adversarial text by original and sampled images. 
Extensive experiments on popular benchmark datasets and VLP models demonstrate that LSSA significantly improves the transferability of multimodal adversarial examples across different V+L downstream tasks.
Additionally, the multimodal adversarial examples crafted by LSSA are more threatening than other attacks on LVLMs.
We hope this work could inspire more works that connects strategies for adversarial attack and defense to evaluate the adversarial robustness of VLP models.

%% file: EMNLP2024/Appendix.tex
\section{More Details}
\subsection{Visualization of Shuffle Images}
\label{Visualization of Shuffle Images}
To better visualize the differences between images after a local shuffle and global shuffle, we present the corner shuffled images in LSSA and the global shuffled images in RLFAT to show the varying degrees of spatial information disruption.
As shown in Figure~\ref{fig:CSSA_shuffle_presentation}, the global shuffled image has been totally disrupted while the corner shuffled image only loses part of the information.
\input{Fig/CSSA_shuffle_presentation}

\subsection{Visualization of Adversarial Examples}
\label{Visualization of Multimodal Adversarial Examples}
\input{Fig/CSSA_adversarial_presentation}
To better show the differences between the generated multimodal adversarial examples and the original samples, we present the image-text pairs and the corresponding adversarial examples generated by LSSA in Figure~\ref{fig:CSSA_adversarial_presentation}.
Due to the small magnitude of the adversarial perturbations, we have amplify the adversarial perturbations by 40 times.

\subsection{Algorithm}
\label{alg_CSSA}
The detailed attack process of our proposed LSSA is described in Algorithm~\ref{alg: CSSA}.
\input{Alg/Algorithm}

\section{More Experiments Details}
\subsection{Experimental Setting}
\label{Experimental Setting}
\textbf{Datasets and VLP Models.}
We conduct experiments on two benchmark datasets, namely Flickr30K~\cite{Flickr30k} and MSCOCO~\cite{MSCOCO}.
Flickr30K and MSCOCO dataset contains 31,783 and 123,287 images, respectively, with each image paired with five corresponding captions.
We evaluate two typical architectures of VLP models: the fused VLP models and aligned models.
For the fused VLP models, we choose ALBEF~\cite{ALBEF} and TCL~\cite{TCL}.
The visual encoder of ALBEF and TCL is ViT-B/16~\cite{ViT}, and the text encoder and multimodal encoder are two 6-layer transformers.
However, they are pre-trained by different pre-trained tasks.
For the aligned VLP models, we choose the CLIP model with different visual encoders.
Specifically, we choose ViT-B/16~\cite{ViT} and ResNet-101~\cite{resnet} as the base visual encoder and BERT as the base text encoder, respectively.

\textbf{Adversarial Attack Settings.}
We follow the multimodal adversarial attack setting of SGA~\cite{SGA}. 
Specifically, we use MI-FGSM attack~\cite{MIM} with perturbation boundary \(\epsilon_v=2/255\), the step size \(\alpha=0.5/255\), and the iteration steps \(T=10\) to generate image adversarial examples.
For text adversarial examples, we adopt BERT-Attack~\cite{bertattack} with the perturbation bound \(\epsilon_t=1\) and the maximum candidate words \(W=10\).
We enlarge the image dataset by resizing the original image as SGA.

\subsection{Performance Metric}
\label{Performance Metric}
We utilize the Attack Success Rate (ASR) as the metric to evaluate the adversarial robustness and transferability of our models in both white-box and black-box settings. 
The ASR measures the percentage of attacks that successfully generate adversarial examples. 
A higher ASR indicates better transferability of adversarial attacks across different models. 
By focusing on ASR, we can assess the effectiveness of our models against adversarial attacks and their ability to generalize beyond the training data, providing a comprehensive evaluation of their robustness and transferability.

\subsection{More Experiments Results}
\label{MSCOCO_main}
\input{Table/CSSA_MSCOCO_main}
We also conduct comprehensive evaluations of our LSSA and other baselines on the MSCOCO dataset.
The experimental settings are consistent with those on the Flickr30K dataset.
As shown in Table~\ref{tab:CSSA_MSCOCO_main}, our LSSA method performs well on two different architecture of models and shows significant improvement compared to the baselines. 
Specifically, LSSA further improves the attack success rate in the white-box setting.
Moreover, in the black-box model setting, it boost the transferable attack performance of multimodal adversarial examples with a clear margin.
Similar to previous observations, adversarial examples generated by models with similar architecture (such as ALBEF and TCL, CLIP\textsubscript{ViT} and CLIP\textsubscript{CNN} ) perform better on other models.

\input{Table/CSSA_eps8_flickr}
We extend the perturbation magnitude of adversarial images and conduct more experiments on the Flickr30K dataset.
As shown in Table~\ref{tab:CSSA_eps_main}, our LSSA method shows significant enhancements compared to other advanced attacks, achieving high attack success rates even on models with different architectures. 
Specifically, in the white-box attack setting, the attack success rate of LSSA nearly approaches 100\%.
In the black-box attack setting, the transferable attack success rate for the ITR task increases by over 15\% even when the model architecture differs.
It indicates that even with larger perturbations, LSSA remains the most advanced transferable attack, demonstrating the robustness of our method.

\subsection{Parameter Study}
In this section, we present the impact of the sampled number of images in adversarial text generation on the transferable attack.
The specific experimental results are shown in the following.

\input{Table/CSSA_text_image_number}
\textbf{Sampled number in text generation.}
We conduct experiments to explore the performance of different sampled numbers around the adversarial images.
As shown in Table~\ref{tab:CSSA_text_image_number}, the increase of sampled number can effectively boost the white-box attack performance and slightly improve the black-box attack success rate, indicating that the performance of adversarial attacks can be enhanced by introducing additional sample information.
To balance between the effectiveness and efficiency, we set the sampled number to 20.

\input{Fig/CSSA_MI}
\textbf{Momentum decay \(\mu\).}
We conduct experiments to investigate the impact of momentum decay.
As shown in Figure~\ref{fig:MI}, the large momentum coefficients \(\mu\) might lead to unstable convergence, while the small one might make the attacker hard to escape from the suboptimal area.
Therefore, the attack performance of LSSA reaches peak when \(\mu=1.0\).

\input{Fig/CSSA_adversarial_perturbation}
\textbf{Sampling boundary \(\epsilon_0\).}
We conduct detailed experiments to further study the impact of sampling boundary \(\epsilon_0\).
As shown in Figure~\ref{fig:epsilon}, the larger sampling boundary results in a slight decrease in attack success rate. 
It may be because the too-large sampling boundary causes the sampled images to be too far from the adversarial image. 
However, the attack success rate does not decrease significantly, possibly because the absolute magnitude of the sampling boundary is not large, keeping the sampling images and the original image within the same area.
Consequently, the differences between the generated adversarial text and the original and sampled image regions remain substantial.

\subsection{Ablation study}
\label{Abaltion study}
\input{Table/CSSA_ablation_study}
To verify the effectiveness of each module, we conduct a series of ablation studies. The following conclusions can be drawn as shown in table~\ref{tab:CSSA_ablation_study}. First, SGA can achieve the desired effect by performing only the first two steps. Furthermore, combined with the sampling module, the attack performance of adversarial examples is boosted on the TCL model, improving 3.48\% (TR R@1) and 2.93\% (IR R@1). On the other hand, integrated with the local shuffle module can also enhance the attack success rate, gaining an improvement of 8.11\% and 5.5\%. After combining the sampling and local shuffle modules, the success rate of white-box attacks also reaches the highest 99.06\% and 98.55\%, and the success rate of black-box attacks increases by 11.48\% and 8.95\% compared with SGA. Finally, momentum further improves the adversarial transferability, and LSSA obtains the best adversarial transferable performance while the while-box attack performance only slightly decays.

%% file: Fig/CSSA_shuffle_presentation.tex
\begin{figure}[h]
\includegraphics[width=0.5\textwidth]{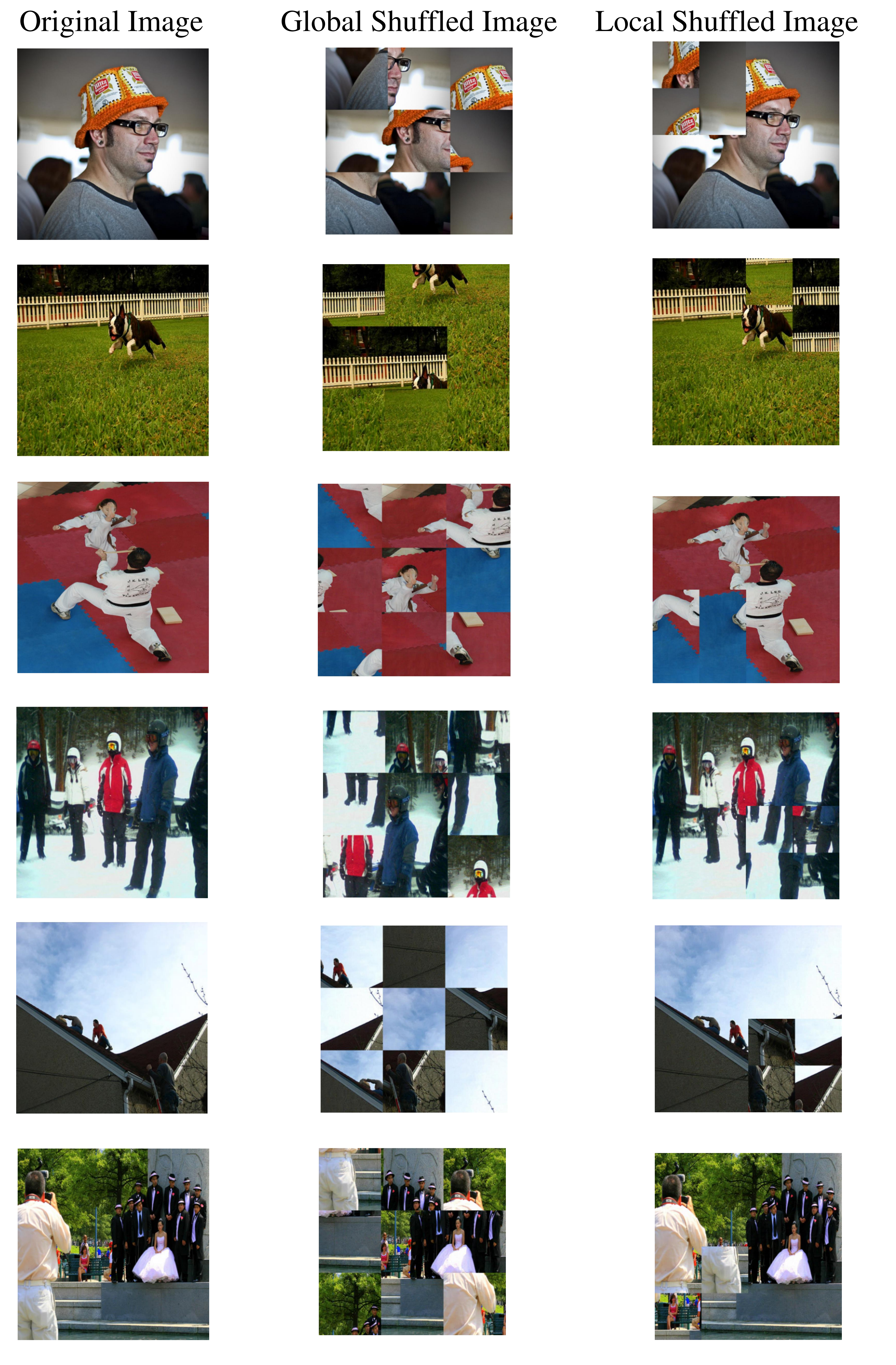}
    \caption{
    Visualization of original images (left), corresponding global shuffle images (middle) and local shuffled images (right). It indicates that local shuffled images preserve more spacial information than global shuffle images while enhancing the diversity.
    }
\label{fig:CSSA_shuffle_presentation}
\end{figure}

%% file: Fig/CSSA_adversarial_presentation.tex
\begin{figure*}[h]
    \includegraphics[width=\textwidth]{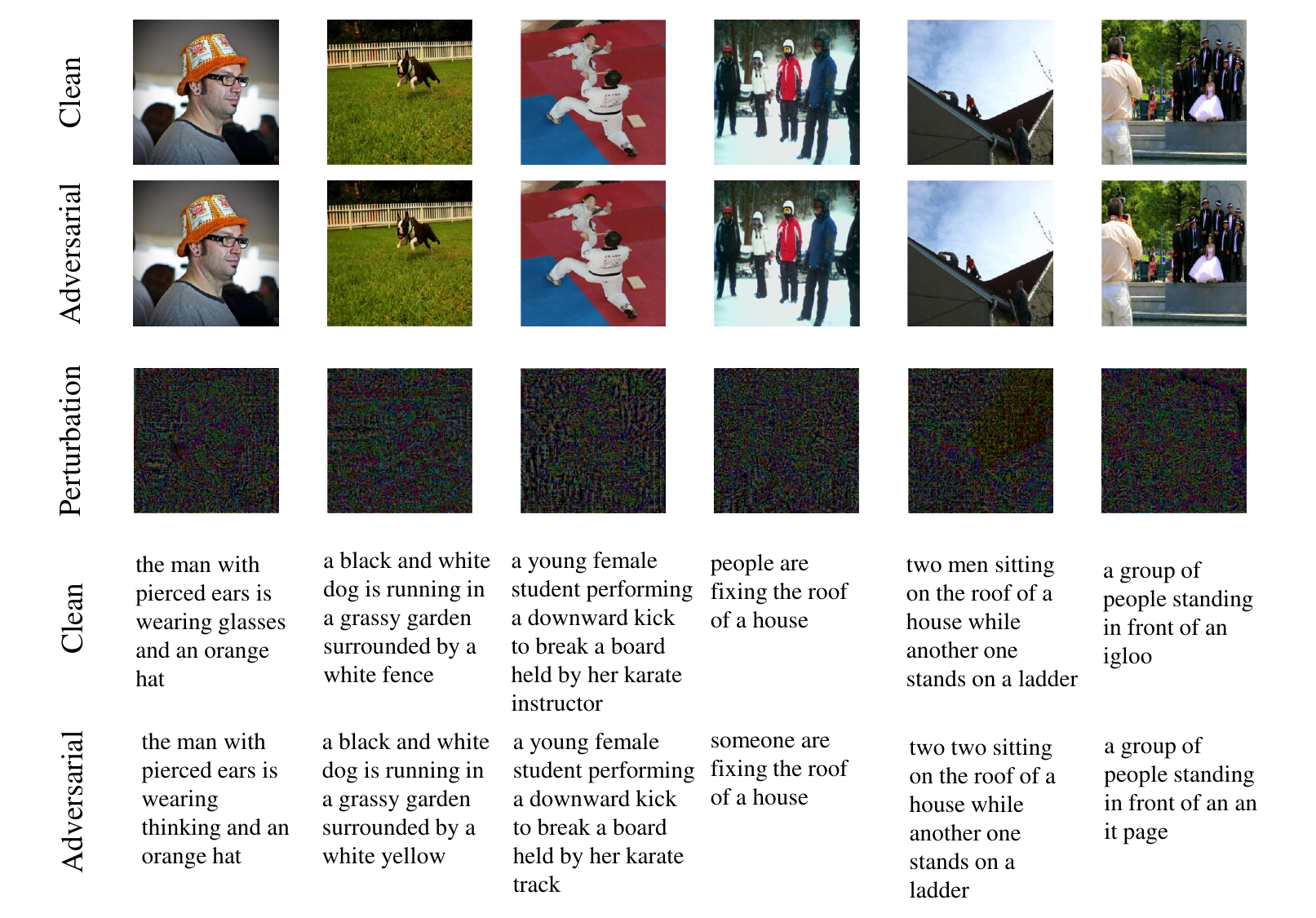}
    \caption{
    Visualization of original and adversarial image-text pairs generated by LSSA. The rows are the original images, corresponding adversarial images and adversarial perturbations, original and corresponding adversarial texts. The perturbation magnitude ensures that the perturbations in the image are imperceptible, and only a few words in the text have changed.
    }
\label{fig:CSSA_adversarial_presentation}
\end{figure*}

%% file: Alg/Algorithm.tex
\begin{algorithm}[h]
    \caption{The Local Shuffle and Sample-based Attack}
    \label{alg: CSSA}
    \begin{algorithmic}
    \State {\bfseries Input:} Image encoder \(f_I\), Text encoder \(f_T\), Dataset \(D\), Image-caption pair \((v,t)\), iteration steps \(T\), random shuffle number \(N\),  sampled number \(M\), loss function \(J\), decay factor \(\lambda\), adversarial step \(\alpha\), image perturbation boundary \(\epsilon_v\), text perturbation boundary \(\epsilon_t\).
    \State \(v^{adv}_1\) \(\leftarrow\) \(v\)
    \State \(g_1\) \(\leftarrow\) \(0\)
    \State // \textbf{Generate adversarial image \(v^{adv}\)}
    \For{$i=1$ {\bfseries to} $T$}
    \State {
    \(g_{i+1}=\lambda \cdot g_{i}+\frac{1}{N}\sum^{N}_{j=1}\frac{\nabla_{v}J(f_I(v^{'}_j),f_T(t))}{\|\nabla_{v}J(f_I(v^{'}_j),f_T(t)\|}\)}
    \State \(v^{adv}_{i+1}\) \(\leftarrow\) \(Clip_{\epsilon_v}(v^{adv}_i+\alpha \cdot sign(g_{i+1}))\)
    \EndFor
    \State \(v^{adv} \leftarrow v^{adv}_{T+1}\)
    \State // \textbf{Generate adversarial text \(t^{adv}\)}
    \State \(t^{adv}=\underset{t^{'}\in B[t,\epsilon_t]}{\arg\max} \quad \lambda \cdot J(f_I(v),f_T(t')) + (1-\lambda)\cdot \frac{1}{M}\sum_{i=1}^{M}J(f_I(v_i),f_T(t'))\)
    \State {\bfseries Output:} adversarial image \(v^{adv}\), adversarial caption \(t^{adv}\)
    \end{algorithmic}
\end{algorithm}

%% file: Table/CSSA_MSCOCO_main.tex
\begin{table*}[t]
\caption{The attack success rate (\%) of multimodal adversarial examples against different VLP models compared with state-of-the-art methods on image-text retrieval task. The source column represents the VLP models used for crafting multimodal adversarial examples on the MSCOCO dataset. * indicates white-box attacks. The higher attack success rate indicates the better performance.
}
\centering
\resizebox{\textwidth}{!}{
\begin{threeparttable}
\begin{tabular}{l|l|cc|cc|cc|cc}
\toprule
\multirow{2}{*}{Source} & \multirow{2}{*}{Attack} &
  \multicolumn{2}{c}{ALBEF} &
  \multicolumn{2}{c}{TCL} &
  \multicolumn{2}{c}{CLIP\textsubscript{ViT}} &
  \multicolumn{2}{c}{CLIP\textsubscript{CNN}} 
  \\
    \cline{3-10}
  & &TR R@1& IR R@1&TR R@1& IR R@1&TR R@1& IR R@1&TR R@1& IR R@1 \\
  \midrule
  \multirow{6}{*}{ALBEF}
&PGD&76.70\textsuperscript{*}&86.30\textsuperscript{*}&12.46&17.77&13.96&23.10&17.45&23.54\\
&BERT-Attack&24.39\textsuperscript{*}&36.13\textsuperscript{*}&24.34&33.39&44.94&52.28&47.73&54.75\\
&Sep-Attack&82.60\textsuperscript{*}&89.88\textsuperscript{*}&32.83&42.92&44.03&54.46&46.96&55.88\\
&Co-Attack&79.87\textsuperscript{*}&87.83\textsuperscript{*}&32.62&43.09&44.89&54.75&47.30&55.64\\
&SGA&96.75\textsuperscript{*}&96.95\textsuperscript{*}&58.56&65.38&57.06&65.25&58.95&66.52\\
&LSSA (Ours)&\textbf{97.89\textsuperscript{*}}&\textbf{97.58\textsuperscript{*}}&\textbf{68.97}&\textbf{73.35}&\textbf{60.85}&\textbf{67.99}&\textbf{65.01}&\textbf{73.33}\\
  \midrule
  \multirow{6}{*}{TCL}
&PGD&10.83&16.52&59.58\textsuperscript{*}&69.53\textsuperscript{*}&14.23&22.28&17.25&23.12\\
&BERT-Attack&35.32&45.92&38.54\textsuperscript{*}&48.48\textsuperscript{*}&51.09&58.80&52.23&61.26\\
&Sep-Attack&41.71&52.97&70.32\textsuperscript{*}&78.97\textsuperscript{*}&50.74&60.13&51.90&61.26\\
&Co-Attack&46.08&57.09&85.38\textsuperscript{*}&91.39\textsuperscript{*}&51.62&60.46&52.13&62.49\\
&SGA&65.93&73.30&98.97\textsuperscript{*}&99.15\textsuperscript{*}&56.34&63.99&59.44&65.70\\
&LSSA (Ours)&\textbf{74.04}&\textbf{80.36}&\textbf{99.23\textsuperscript{*}}&\textbf{99.33\textsuperscript{*}}&\textbf{60.70}&\textbf{67.43}&\textbf{62.40}&\textbf{69.63}\\
  \midrule
  \multirow{6}{*}{CLIP\textsubscript{ViT}}
&PGD&~~7.24&10.75&10.19&13.74&54.79\textsuperscript{*}&66.85\textsuperscript{*}&~~7.32&11.34\\
&BERT-Attack&20.34&29.74&21.08&29.61&45.06\textsuperscript{*}&51.68\textsuperscript{*}&44.54&53.72\\
&Sep-Attack&23.41&34.61&25.77&36.84&68.52\textsuperscript{*}&77.94\textsuperscript{*}&43.11&49.76\\
&Co-Attack&30.28&42.67&32.84&44.69&97.98\textsuperscript{*}&98.80\textsuperscript{*}&55.08&62.51\\
&SGA&33.41&44.64&37.54&47.76&99.69\textsuperscript{*}&99.69\textsuperscript{*}&58.93&65.83\\
&LSSA (Ours)&\textbf{38.88}&\textbf{49.66}&\textbf{40.63}&\textbf{52.95}&\textbf{99.72\textsuperscript{*}}&\textbf{99.77\textsuperscript{*}}&\textbf{69.02}&\textbf{73.17}\\
  \midrule
  \multirow{6}{*}{CLIP\textsubscript{CNN}}
&PGD&~~7.01&10.62&10.08&13.65&~~4.88&10.70&76.99\textsuperscript{*}&84.20\textsuperscript{*}\\
&BERT-Attack&23.38&34.64&24.58&29.61&51.28&57.49&54.43\textsuperscript{*}&62.17\textsuperscript{*}\\
&Sep-Attack&26.53&39.29&30.26&41.51&50.44&57.11&88.72\textsuperscript{*}&92.49\textsuperscript{*}\\
&Co-Attack&29.83&41.97&32.97&43.72&53.10&58.90&96.72\textsuperscript{*}&98.56\textsuperscript{*}\\
&SGA&31.61&43.00&34.81&45.95&56.62&60.77&99.61\textsuperscript{*}&99.80\textsuperscript{*}\\
&LSSA (Ours)&\textbf{37.92}&\textbf{47.90}&\textbf{34.81}&\textbf{47.18}&\textbf{61.30}&\textbf{68.68}&\textbf{99.88\textsuperscript{*}}&\textbf{99.99\textsuperscript{*}}\\
\bottomrule
\end{tabular}
\end{threeparttable}
}
\label{tab:CSSA_MSCOCO_main}
\end{table*}

%% file: Table/CSSA_eps8_flickr.tex
\begin{table*}[t]
\caption{The attack success rate (\%) of multimodal adversarial examples with perturbation boundary \(\epsilon_v=8/255\) against different VLP models compared with state-of-the-art methods on image-text retrieval task. The source column represents the VLP models used for crafting multimodal adversarial examples on Flickr30K dataset. * indicates white-box attacks. The higher attack success rate indicates the better performance.
}
\centering
\resizebox{\textwidth}{!}{
\begin{threeparttable}
\begin{tabular}{l|l|cc|cc|cc|cc}
\toprule
\multirow{2}{*}{Source} & \multirow{2}{*}{Attack} &
  \multicolumn{2}{c}{ALBEF} &
  \multicolumn{2}{c}{TCL} &
  \multicolumn{2}{c}{CLIP\textsubscript{ViT}} &
  \multicolumn{2}{c}{CLIP\textsubscript{CNN}} 
  \\
    \cline{3-10}
  & &TR R@1& IR R@1&TR R@1& IR R@1&TR R@1& IR R@1&TR R@1& IR R@1 \\
  \midrule
  \multirow{6}{*}{ALBEF}
&PGD&93.74\textsuperscript{*}&94.43\textsuperscript{*}&24.03&27.9&10.67&15.82&14.05&19.11\\
&BERT-Attack&11.57\textsuperscript{*}&27.46\textsuperscript{*}&12.64&28.07&29.33&43.17&32.69&46.11\\
&Sep-Attack&95.72\textsuperscript{*}&96.14\textsuperscript{*}&39.30&51.79&34.11&45.72&35.76&47.92\\
&Co-Attack&97.08\textsuperscript{*}&98.36\textsuperscript{*}&39.52&51.24&29.82&38.92&31.29&41.99\\
&SGA&99.79\textsuperscript{*}&99.95\textsuperscript{*}&88.30&87.81&37.30&46.17&39.85&50.67\\
&LSSA (Ours)&\textbf{99.90\textsuperscript{*}}&\textbf{99.98\textsuperscript{*}}&\textbf{96.00}&\textbf{96.07}&\textbf{54.36}&\textbf{60.60}&\textbf{54.79}&\textbf{64.63}\\
  \midrule
  \multirow{6}{*}{TCL}
&PGD&35.77&41.67&99.37\textsuperscript{*}&99.33\textsuperscript{*}&10.18&16.30&14.81&21.10\\
&BERT-Attack&11.89&26.82&14.54\textsuperscript{*}&29.17&29.69&44.49&33.46&46.07\\
&Sep-Attack&52.45&61.44&99.58\textsuperscript{*}&99.45\textsuperscript{*}&37.06&45.81&37.42&49.91\\
&Co-Attack&49.84&60.36&91.68\textsuperscript{*}&95.48\textsuperscript{*}&32.64&42.69&32.06&47.82\\
&SGA&93.12&92.89&98.42\textsuperscript{*}&98.76\textsuperscript{*}&35.34&45.75&40.10&50.22\\
&LSSA (Ours)&\textbf{98.23}&\textbf{98.50}&\textbf{100.0\textsuperscript{*}}&\textbf{100.0\textsuperscript{*}}&\textbf{53.50}&\textbf{60.60}&\textbf{56.45}&\textbf{66.52}\\
  \midrule
  \multirow{6}{*}{CLIP\textsubscript{ViT}}
&PGD&~~3.13&~~6.48&~~4.43&~~8.83&69.33\textsuperscript{*}&84.79\textsuperscript{*}&13.03&17.43\\
&BERT-Attack&~~9.59&22.64&11.80&25.07&28.34\textsuperscript{*}&39.08\textsuperscript{*}&30.40&37.43\\
&Sep-Attack&~~7.61&20.58&10.12&20.74&76.93\textsuperscript{*}&87.44\textsuperscript{*}&29.89&38.32\\
&Co-Attack&~~8.55&20.18&10.01&21.29&78.53\textsuperscript{*}&87.50\textsuperscript{*}&29.50&38.49\\
&SGA&22.63&35.15&26.55&37.26&99.26\textsuperscript{*}&99.10\textsuperscript{*}&54.92&61.41\\
&LSSA (Ours)&\textbf{45.88}&\textbf{57.18}&\textbf{45.10}&\textbf{56.57}&\textbf{100.0\textsuperscript{*}}&\textbf{100.0\textsuperscript{*}}&\textbf{77.27}&\textbf{80.96}\\
  \midrule
  \multirow{6}{*}{CLIP\textsubscript{CNN}}
&PGD&~~2.29&~~6.15&~~4.53&~~8.88&~~5.40&12.08&89.78\textsuperscript{*}&93.04\textsuperscript{*}\\
&BERT-Attack&~~8.86&23.27&12.33&25.48&27.12&37.44&30.40\textsuperscript{*}&40.10\textsuperscript{*}\\
&Sep-Attack&~~9.38&22.99&11.28&25.45&26.13&39.24&93.61\textsuperscript{*}&95.30\textsuperscript{*}\\
&Co-Attack&10.53&23.62&12.54&26.05&27.24&40.62&95.91\textsuperscript{*}&96.50\textsuperscript{*}\\
&SGA&16.37&28.74&18.76&33.14&38.52&52.00&99.11\textsuperscript{*}&99.49\textsuperscript{*}\\
&LSSA (Ours)&\textbf{31.49}&\textbf{43.92}&\textbf{35.72}&\textbf{47.19}&\textbf{64.17}&\textbf{69.10}&\textbf{100.0\textsuperscript{*}}&\textbf{100.0\textsuperscript{*}}\\
\bottomrule
\end{tabular}
\end{threeparttable}
}
\label{tab:CSSA_eps_main}
\end{table*}

%% file: Table/CSSA_text_image_number.tex
\begin{table*}[t]
\caption{The attack success rate (\%) of adversarial text generation with different sampled numbers against different VLP models. The source column represents the VLP models used for crafting multimodal adversarial examples on the Flickr30K dataset. * indicates white-box attacks. The higher attack success rate indicates the better performance.
}
\centering
\resizebox{\textwidth}{!}{
\begin{threeparttable}
\begin{tabular}{l|c|cc|cc|cc|cc}
\toprule
\multirow{2}{*}{Source} & \multirow{2}{*}{Number} &
  \multicolumn{2}{c}{ALBEF} &
  \multicolumn{2}{c}{TCL} &
  \multicolumn{2}{c}{CLIP\textsubscript{ViT}} &
  \multicolumn{2}{c}{CLIP\textsubscript{CNN}} 
  \\
    \cline{3-10}
  & &TR R@1& IR R@1&TR R@1& IR R@1&TR R@1& IR R@1&TR R@1& IR R@1 \\
  \midrule
  \multirow{8}{*}{ALBEF}
&0&96.45\textsuperscript{*}&96.40\textsuperscript{*}&56.16&64.38&35.95&48.16&40.38&50.23\\
&1&97.50\textsuperscript{*}&97.10\textsuperscript{*}&59.33&65.88&35.83&48.13&40.45&50.33\\
&2&97.50\textsuperscript{*}&97.10\textsuperscript{*}&59.33&65.90&35.83&48.20&40.45&50.33\\
&5&97.50\textsuperscript{*}&97.10\textsuperscript{*}&59.33&65.88&35.98&48.22&40.47&50.52\\
&10&97.50\textsuperscript{*}&97.10\textsuperscript{*}&59.33&65.90&35.98&48.22&40.87&50.52\\
&20&97.50\textsuperscript{*}&97.10\textsuperscript{*}&59.33&65.88&36.32&48.49&40.87&50.77\\
&30&\textbf{97.50}\textsuperscript{*}&\textbf{97.10}\textsuperscript{*}&\textbf{59.33}&\textbf{65.90}&\textbf{36.32}&\textbf{48.49}&\textbf{40.91}&\textbf{50.80}\\
\bottomrule
\end{tabular}
\end{threeparttable}
}
\label{tab:CSSA_text_image_number}
\end{table*}

%% file: Fig/CSSA_MI.tex
\begin{figure}[h]
    \centering
    \includegraphics[width=\columnwidth]{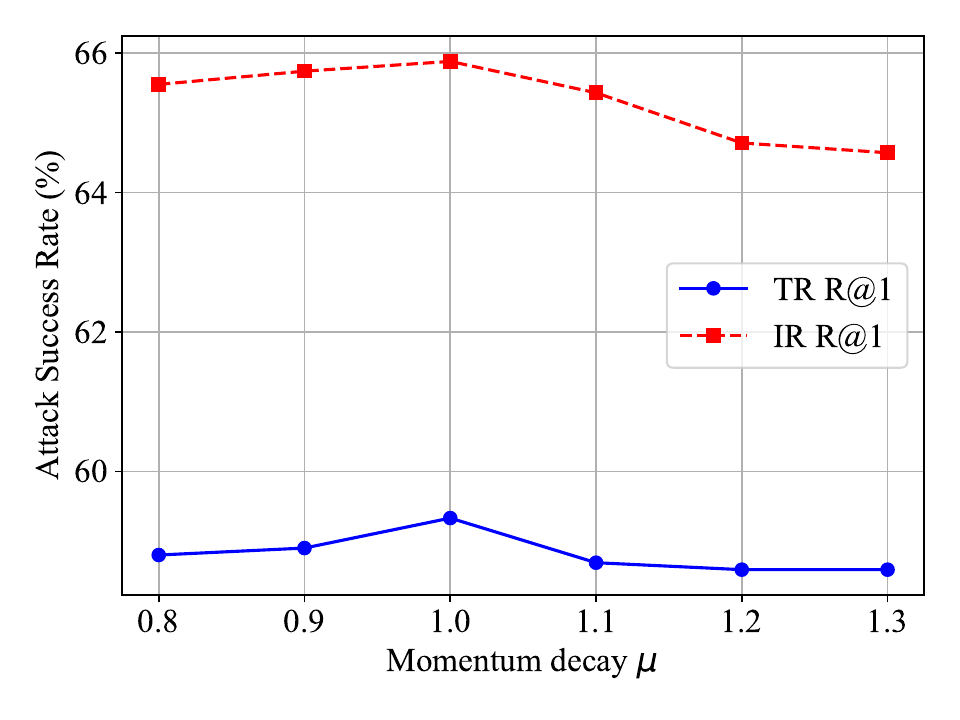}
    \caption{ 
    The attack success rate (\%) of LSSA with different momentum decay \(\mu\) on the Flickr30K dataset. The source and target model are ALBEF and TCL model, respectively.
    }
    \label{fig:MI}
\end{figure}

%% file: Fig/CSSA_adversarial_perturbation.tex
\begin{figure}[h]
    \centering
    \includegraphics[width=\columnwidth]{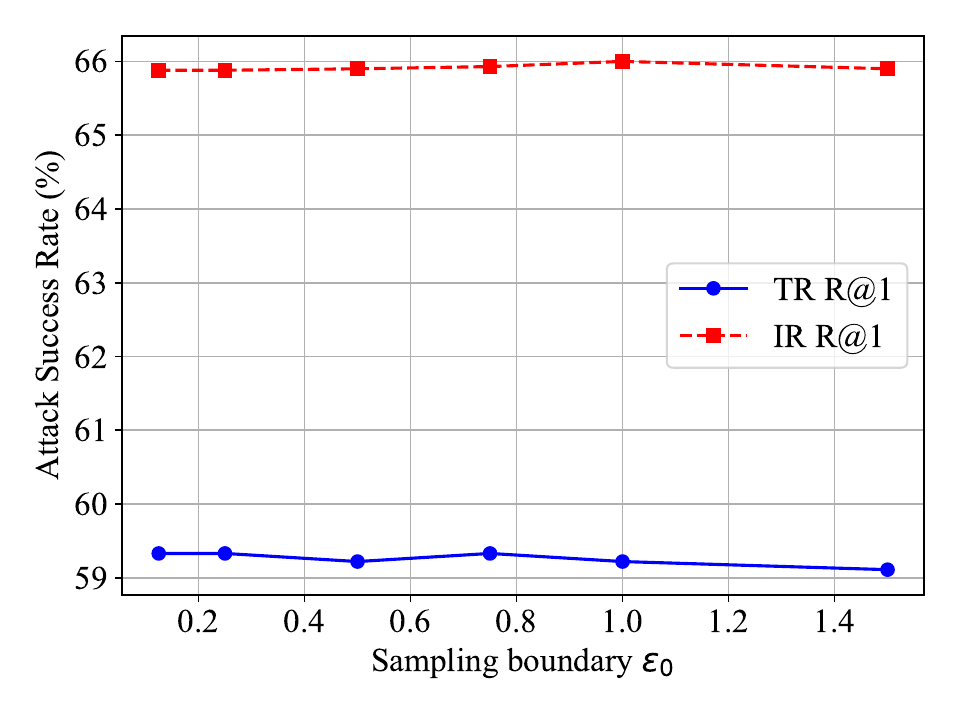}
    \caption{ 
    The attack success rate (\%) of LSSA with different sample magnitude \(\epsilon\) on the Flickr30K dataset. The source and target model are ALBEF and TCL model, respectively.
    }
    \label{fig:epsilon}
\end{figure}

%% file: Table/CSSA_ablation_study.tex
\begin{table*}[t]
\caption{Ablation study of the proposed LSSA. SGA$_{2}$ indicates a variant of SGA which performs only the first two steps. SA and LS denote the sampling and local shuffle modules, respectively. The difference between SGA\textsubscript{2}+SA+LS and LSSA is the former does not contain momentum while the latter contains. ALBEF is used as the white-box model to craft multimodal adversarial examples on the Flickr30K dataset. * indicates white-box attacks. The higher attack success rate indicates the better performance.}
\resizebox{\textwidth}{!}{
\begin{threeparttable}
\begin{tabular}{l|l|cccccc|cccccc}
\toprule
\multirow{2}{*}{Source} & \multirow{2}{*}{Attack} &
  \multicolumn{6}{c}{ALBEF} &
  \multicolumn{6}{c}{TCL} 
  \\
    \cline{3-14}
  & &TR R@1&TR R@5&TR R@10& IR R@1&IR R@5&IR R@10&TR R@1&TR R@5&TR R@10& IR R@1&IR R@5&IR R@10 \\
\midrule
\multirow{6}{*}{ALBEF}
& SGA & 97.24\textsuperscript{*}&94.59\textsuperscript{*}&92.20\textsuperscript{*}&97.08\textsuperscript{*}&94.30\textsuperscript{*}&92.38\textsuperscript{*}&44.68 &24.02&16.13&55.57&36.01&27.60\\
& SGA\textsubscript{2} &96.14\textsuperscript{*}&92.69\textsuperscript{*}&90.20\textsuperscript{*}&96.66\textsuperscript{*}&93.31\textsuperscript{*}&91.41\textsuperscript{*}&45.42&22.41&15.53&56.90&37.34&29.14 \\
& SGA\textsubscript{2}$+$SA & 97.81\textsuperscript{*}&95.89\textsuperscript{*}&93.80\textsuperscript{*}&97.26\textsuperscript{*}&94.67\textsuperscript{*}&92.98\textsuperscript{*}&48.16&26.53&18.84&58.52&38.12&29.77\\
& SGA\textsubscript{2}$+$LS &98.12\textsuperscript{*}&95.89\textsuperscript{*}&94.46\textsuperscript{*}&98.22\textsuperscript{*}&96.37 \textsuperscript{*}&95.13\textsuperscript{*}&52.79&31.16&23.85&61.07&41.68&32.86\\
& SGA\textsubscript{2}$+$SA$+$LS & \textbf{99.06}\textsuperscript{*}&\textbf{97.70}\textsuperscript{*}&\textbf{96.30}\textsuperscript{*}&\textbf{98.55}\textsuperscript{*}&\textbf{96.86}\textsuperscript{*}&\textbf{95.67}\textsuperscript{*}&56.16&35.18&26.25&64.52&45.34&36.21\\
& LSSA & 97.29\textsuperscript{*}&94.59\textsuperscript{*}&92.50\textsuperscript{*}&97.22\textsuperscript{*}&94.09\textsuperscript{*}&91.99\textsuperscript{*}&\textbf{59.43}&\textbf{36.58}&\textbf{26.45}&\textbf{65.62}&\textbf{47.14}&\textbf{38.40}\\
\bottomrule
\end{tabular}
\end{threeparttable}
}
\label{tab:CSSA_ablation_study}
\end{table*}

%% file: main.bbl
\begin{thebibliography}{42}
\providecommand{\natexlab}[1]{#1}

\bibitem[{Anderson et~al.(2016)Anderson, Fernando, Johnson, and Gould}]{SPICE}
Peter Anderson, Basura Fernando, Mark Johnson, and Stephen Gould. 2016.
\newblock {SPICE:} semantic propositional image caption evaluation.
\newblock In \emph{{ECCV}}, pages 382--398.

\bibitem[{Banerjee and Lavie(2005)}]{METEOR}
Satanjeev Banerjee and Alon Lavie. 2005.
\newblock {METEOR:} an automatic metric for {MT} evaluation with improved correlation with human judgments.
\newblock In \emph{Proceedings of the Workshop on Intrinsic and Extrinsic Evaluation Measures for Machine Translation and/or Summarization@ACL 2005, Ann Arbor, Michigan, USA, June 29, 2005}, pages 65--72.

\bibitem[{Cao et~al.(2022)Cao, Li, Li, Nie, and Zhang}]{ITRsurvey}
Min Cao, Shiping Li, Juntao Li, Liqiang Nie, and Min Zhang. 2022.
\newblock Image-text retrieval: A survey on recent research and development.
\newblock \emph{arXiv preprint arXiv:2203.14713}.

\bibitem[{Chen et~al.(2020{\natexlab{a}})Chen, Ding, Liu, Lin, Liu, and Han}]{ITR}
Hui Chen, Guiguang Ding, Xudong Liu, Zijia Lin, Ji~Liu, and Jungong Han. 2020{\natexlab{a}}.
\newblock {IMRAM:} iterative matching with recurrent attention memory for cross-modal image-text retrieval.
\newblock In \emph{{CVPR}}, pages 12652--12660.

\bibitem[{Chen et~al.(2020{\natexlab{b}})Chen, Li, Yu, Kholy, Ahmed, Gan, Cheng, and Liu}]{UNITER}
Yen{-}Chun Chen, Linjie Li, Licheng Yu, Ahmed~El Kholy, Faisal Ahmed, Zhe Gan, Yu~Cheng, and Jingjing Liu. 2020{\natexlab{b}}.
\newblock {UNITER:} universal image-text representation learning.
\newblock In \emph{{ECCV}}, pages 104--120.

\bibitem[{Dong et~al.(2018)Dong, Liao, Pang, Su, Zhu, Hu, and Li}]{MIM}
Yinpeng Dong, Fangzhou Liao, Tianyu Pang, Hang Su, Jun Zhu, Xiaolin Hu, and Jianguo Li. 2018.
\newblock Boosting adversarial attacks with momentum.
\newblock In \emph{{CVPR}}, pages 9185--9193.

\bibitem[{Dong et~al.(2019)Dong, Pang, Su, and Zhu}]{TIM}
Yinpeng Dong, Tianyu Pang, Hang Su, and Jun Zhu. 2019.
\newblock Evading defenses to transferable adversarial examples by translation-invariant attacks.
\newblock In \emph{{CVPR}}, pages 4312--4321.

\bibitem[{Dosovitskiy et~al.(2021)Dosovitskiy, Beyer, Kolesnikov, Weissenborn, Zhai, Unterthiner, Dehghani, Minderer, Heigold, Gelly, Uszkoreit, and Houlsby}]{ViT}
Alexey Dosovitskiy, Lucas Beyer, Alexander Kolesnikov, Dirk Weissenborn, Xiaohua Zhai, Thomas Unterthiner, Mostafa Dehghani, Matthias Minderer, Georg Heigold, Sylvain Gelly, Jakob Uszkoreit, and Neil Houlsby. 2021.
\newblock An image is worth 16x16 words: Transformers for image recognition at scale.
\newblock In \emph{{ICLR}}.

\bibitem[{Dou et~al.(2022)Dou, Xu, Gan, Wang, Wang, Wang, Zhu, Zhang, Yuan, Peng, Liu, and Zeng}]{DBLP:conf/cvpr/DouXGWWWZZYP0022}
Zi{-}Yi Dou, Yichong Xu, Zhe Gan, Jianfeng Wang, Shuohang Wang, Lijuan Wang, Chenguang Zhu, Pengchuan Zhang, Lu~Yuan, Nanyun Peng, Zicheng Liu, and Michael Zeng. 2022.
\newblock An empirical study of training end-to-end vision-and-language transformers.
\newblock In \emph{{CVPR}}, pages 18145--18155.

\bibitem[{Du et~al.(2022)Du, Qian, Liu, Ding, Qiu, Yang, and Tang}]{visualGLM}
Zhengxiao Du, Yujie Qian, Xiao Liu, Ming Ding, Jiezhong Qiu, Zhilin Yang, and Jie Tang. 2022.
\newblock {GLM:} general language model pretraining with autoregressive blank infilling.
\newblock In \emph{{ACL}}, pages 320--335.

\bibitem[{Goodfellow et~al.(2015)Goodfellow, Shlens, and Szegedy}]{fgsm}
Ian~J. Goodfellow, Jonathon Shlens, and Christian Szegedy. 2015.
\newblock Explaining and harnessing adversarial examples.
\newblock In \emph{3rd International Conference on Learning Representations, {ICLR}}.

\bibitem[{He et~al.(2016)He, Zhang, Ren, and Sun}]{resnet}
Kaiming He, Xiangyu Zhang, Shaoqing Ren, and Jian Sun. 2016.
\newblock Deep residual learning for image recognition.
\newblock In \emph{{CVPR}}, pages 770--778.

\bibitem[{Li et~al.(2023{\natexlab{a}})Li, Li, Savarese, and Hoi}]{BLIP2}
Junnan Li, Dongxu Li, Silvio Savarese, and Steven C.~H. Hoi. 2023{\natexlab{a}}.
\newblock {BLIP-2:} bootstrapping language-image pre-training with frozen image encoders and large language models.
\newblock In \emph{{ICML}}, pages 19730--19742.

\bibitem[{Li et~al.(2022)Li, Li, Xiong, and Hoi}]{BLIP}
Junnan Li, Dongxu Li, Caiming Xiong, and Steven C.~H. Hoi. 2022.
\newblock {BLIP:} bootstrapping language-image pre-training for unified vision-language understanding and generation.
\newblock In \emph{{ICML}}, pages 12888--12900.

\bibitem[{Li et~al.(2021)Li, Selvaraju, Gotmare, Joty, Xiong, and Hoi}]{ALBEF}
Junnan Li, Ramprasaath~R. Selvaraju, Akhilesh Gotmare, Shafiq~R. Joty, Caiming Xiong, and Steven~Chu{-}Hong Hoi. 2021.
\newblock Align before fuse: Vision and language representation learning with momentum distillation.
\newblock In \emph{{NeurIPS}}, pages 9694--9705.

\bibitem[{Li et~al.(2020{\natexlab{a}})Li, Ma, Guo, Xue, and Qiu}]{bertattack}
Linyang Li, Ruotian Ma, Qipeng Guo, Xiangyang Xue, and Xipeng Qiu. 2020{\natexlab{a}}.
\newblock {BERT-ATTACK:} adversarial attack against {BERT} using {BERT}.
\newblock In \emph{EMNLP}, pages 6193--6202.

\bibitem[{Li et~al.(2023{\natexlab{b}})Li, Guo, Zuo, and Chen}]{STAT}
Qizhang Li, Yiwen Guo, Wangmeng Zuo, and Hao Chen. 2023{\natexlab{b}}.
\newblock Squeeze training for adversarial robustness.
\newblock In \emph{{ICLR}}.

\bibitem[{Li et~al.(2020{\natexlab{b}})Li, Yin, Li, Zhang, Hu, Zhang, Wang, Hu, Dong, Wei, Choi, and Gao}]{Oscar}
Xiujun Li, Xi~Yin, Chunyuan Li, Pengchuan Zhang, Xiaowei Hu, Lei Zhang, Lijuan Wang, Houdong Hu, Li~Dong, Furu Wei, Yejin Choi, and Jianfeng Gao. 2020{\natexlab{b}}.
\newblock Oscar: Object-semantics aligned pre-training for vision-language tasks.
\newblock In \emph{{ECCV}}, pages 121--137.

\bibitem[{Lin(2004)}]{ROUGE}
Chin-Yew Lin. 2004.
\newblock Rouge: A package for automatic evaluation of summaries.
\newblock In \emph{Text summarization branches out}, pages 74--81.

\bibitem[{Lin et~al.(2020)Lin, Song, He, Wang, and Hopcroft}]{NI}
Jiadong Lin, Chuanbiao Song, Kun He, Liwei Wang, and John~E. Hopcroft. 2020.
\newblock Nesterov accelerated gradient and scale invariance for adversarial attacks.
\newblock In \emph{{ICLR}}.

\bibitem[{Lin et~al.(2014)Lin, Maire, Belongie, Hays, Perona, Ramanan, Doll{\'{a}}r, and Zitnick}]{MSCOCO}
Tsung{-}Yi Lin, Michael Maire, Serge~J. Belongie, James Hays, Pietro Perona, Deva Ramanan, Piotr Doll{\'{a}}r, and C.~Lawrence Zitnick. 2014.
\newblock Microsoft {COCO:} common objects in context.
\newblock In \emph{{ECCV}}, pages 740--755.

\bibitem[{Lu et~al.(2023)Lu, Wang, Wang, Guan, Gao, and Zheng}]{SGA}
Dong Lu, Zhiqiang Wang, Teng Wang, Weili Guan, Hongchang Gao, and Feng Zheng. 2023.
\newblock Set-level guidance attack: Boosting adversarial transferability of vision-language pre-training models.
\newblock In \emph{{ICCV}}, pages 102--111.

\bibitem[{Papineni et~al.(2002)Papineni, Roukos, Ward, and Zhu}]{BLEU}
Kishore Papineni, Salim Roukos, Todd Ward, and Wei{-}Jing Zhu. 2002.
\newblock Bleu: a method for automatic evaluation of machine translation.
\newblock In \emph{{ACL}}, pages 311--318.

\bibitem[{Plummer et~al.(2017)Plummer, Wang, Cervantes, Caicedo, Hockenmaier, and Lazebnik}]{Flickr30k}
Bryan~A. Plummer, Liwei Wang, Chris~M. Cervantes, Juan~C. Caicedo, Julia Hockenmaier, and Svetlana Lazebnik. 2017.
\newblock Flickr30k entities: Collecting region-to-phrase correspondences for richer image-to-sentence models.
\newblock \emph{Int. J. Comput. Vis.}, 123(1):74--93.

\bibitem[{Radford et~al.(2021)Radford, Kim, Hallacy, Ramesh, Goh, Agarwal, Sastry, Askell, Mishkin, Clark, Krueger, and Sutskever}]{CLIP}
Alec Radford, Jong~Wook Kim, Chris Hallacy, Aditya Ramesh, Gabriel Goh, Sandhini Agarwal, Girish Sastry, Amanda Askell, Pamela Mishkin, Jack Clark, Gretchen Krueger, and Ilya Sutskever. 2021.
\newblock Learning transferable visual models from natural language supervision.
\newblock In \emph{{ICML}}, pages 8748--8763.

\bibitem[{Sadhu et~al.(2019)Sadhu, Chen, and Nevatia}]{VG}
Arka Sadhu, Kan Chen, and Ram Nevatia. 2019.
\newblock Zero-shot grounding of objects from natural language queries.
\newblock In \emph{{ICCV}}, pages 4693--4702.

\bibitem[{Song et~al.(2020)Song, He, Lin, Wang, and Hopcroft}]{RLFAT}
Chuanbiao Song, Kun He, Jiadong Lin, Liwei Wang, and John~E. Hopcroft. 2020.
\newblock Robust local features for improving the generalization of adversarial training.
\newblock In \emph{{ICLR}}.

\bibitem[{Su et~al.(2023)Su, Lan, Li, Xu, Wang, and Cai}]{PandaGPT}
Yixuan Su, Tian Lan, Huayang Li, Jialu Xu, Yan Wang, and Deng Cai. 2023.
\newblock Pandagpt: One model to instruction-follow them all.
\newblock \emph{CoRR}, abs/2305.16355.

\bibitem[{Touvron et~al.(2021)Touvron, Cord, Douze, Massa, Sablayrolles, and J{\'{e}}gou}]{DBLP:conf/icml/TouvronCDMSJ21}
Hugo Touvron, Matthieu Cord, Matthijs Douze, Francisco Massa, Alexandre Sablayrolles, and Herv{\'{e}} J{\'{e}}gou. 2021.
\newblock Training data-efficient image transformers {\&} distillation through attention.
\newblock In \emph{{ICML}}, pages 10347--10357.

\bibitem[{Vedantam et~al.(2015)Vedantam, Zitnick, and Parikh}]{CIDEr}
Ramakrishna Vedantam, C.~Lawrence Zitnick, and Devi Parikh. 2015.
\newblock Cider: Consensus-based image description evaluation.
\newblock In \emph{{CVPR}}, pages 4566--4575.

\bibitem[{Vinyals et~al.(2015)Vinyals, Toshev, Bengio, and Erhan}]{IC}
Oriol Vinyals, Alexander Toshev, Samy Bengio, and Dumitru Erhan. 2015.
\newblock Show and tell: {A} neural image caption generator.
\newblock In \emph{{CVPR}}, pages 3156--3164.

\bibitem[{Wang et~al.(2023)Wang, Ge, Zheng, Cheng, Shan, Qie, and Luo}]{DBLP:conf/cvpr/WangGZCSQL23}
Teng Wang, Yixiao Ge, Feng Zheng, Ran Cheng, Ying Shan, Xiaohu Qie, and Ping Luo. 2023.
\newblock Accelerating vision-language pretraining with free language modeling.
\newblock In \emph{{CVPR}}, pages 23161--23170.

\bibitem[{Wang et~al.(2022)Wang, Jiang, Lu, Zheng, Cheng, Yin, and Luo}]{VLMixer}
Teng Wang, Wenhao Jiang, Zhichao Lu, Feng Zheng, Ran Cheng, Chengguo Yin, and Ping Luo. 2022.
\newblock Vlmixer: Unpaired vision-language pre-training via cross-modal cutmix.
\newblock In \emph{{ICML}}, pages 22680--22690.

\bibitem[{Wang et~al.(2021)Wang, Yang, Deng, and He}]{FGPM}
Xiaosen Wang, Yichen Yang, Yihe Deng, and Kun He. 2021.
\newblock Adversarial training with fast gradient projection method against synonym substitution based text attacks.
\newblock In \emph{{AAAI}}, pages 13997--14005.

\bibitem[{Wang et~al.(2019)Wang, Liu, Li, Sheng, Yan, Wang, and Shao}]{TIR}
Zihao Wang, Xihui Liu, Hongsheng Li, Lu~Sheng, Junjie Yan, Xiaogang Wang, and Jing Shao. 2019.
\newblock {CAMP:} cross-modal adaptive message passing for text-image retrieval.
\newblock In \emph{{ICCV}}, pages 5763--5772.

\bibitem[{Xie et~al.(2019)Xie, Zhang, Zhou, Bai, Wang, Ren, and Yuille}]{DIM}
Cihang Xie, Zhishuai Zhang, Yuyin Zhou, Song Bai, Jianyu Wang, Zhou Ren, and Alan~L. Yuille. 2019.
\newblock Improving transferability of adversarial examples with input diversity.
\newblock In \emph{{CVPR}}, pages 2730--2739.

\bibitem[{Yang et~al.(2022)Yang, Duan, Tran, Xu, Chanda, Chen, Zeng, Chilimbi, and Huang}]{TCL}
Jinyu Yang, Jiali Duan, Son Tran, Yi~Xu, Sampath Chanda, Liqun Chen, Belinda Zeng, Trishul Chilimbi, and Junzhou Huang. 2022.
\newblock Vision-language pre-training with triple contrastive learning.
\newblock In \emph{{CVPR}}, pages 15650--15659.

\bibitem[{Yu et~al.(2023)Yu, Qin, Chen, Lian, Fu, Wen, Xue, and He}]{Sparse}
Zhen Yu, Zhou Qin, Zhenhua Chen, Meihui Lian, Haojun Fu, Weigao Wen, Hui Xue, and Kun He. 2023.
\newblock Sparse black-box multimodal attack for vision-language adversary generation.
\newblock In \emph{{EMNLP}}, pages 5775--5784.

\bibitem[{Yuan et~al.(2021)Yuan, Chen, Wang, Yu, Shi, Jiang, Tay, Feng, and Yan}]{DBLP:conf/iccv/0007CWYSJTFY21}
Li~Yuan, Yunpeng Chen, Tao Wang, Weihao Yu, Yujun Shi, Zihang Jiang, Francis E.~H. Tay, Jiashi Feng, and Shuicheng Yan. 2021.
\newblock Tokens-to-token vit: Training vision transformers from scratch on imagenet.
\newblock In \emph{{ICCV}}, pages 538--547.

\bibitem[{Zhang et~al.(2022)Zhang, Yi, and Sang}]{Co-attack}
Jiaming Zhang, Qi~Yi, and Jitao Sang. 2022.
\newblock Towards adversarial attack on vision-language pre-training models.
\newblock In \emph{{ACM MM}}, pages 5005--5013.

\bibitem[{Zhang et~al.(2021)Zhang, Li, Hu, Yang, Zhang, Wang, Choi, and Gao}]{VinVL}
Pengchuan Zhang, Xiujun Li, Xiaowei Hu, Jianwei Yang, Lei Zhang, Lijuan Wang, Yejin Choi, and Jianfeng Gao. 2021.
\newblock Vinvl: Revisiting visual representations in vision-language models.
\newblock In \emph{{CVPR}}, pages 5579--5588.

\bibitem[{Zhu et~al.(2023)Zhu, Chen, Shen, Li, and Elhoseiny}]{MiniGPT-4}
Deyao Zhu, Jun Chen, Xiaoqian Shen, Xiang Li, and Mohamed Elhoseiny. 2023.
\newblock Minigpt-4: Enhancing vision-language understanding with advanced large language models.
\newblock \emph{CoRR}, abs/2304.10592.

\end{thebibliography}
